\documentclass[10pt,twocolumn,letterpaper]{article}

\usepackage[pagenumbers]{cvpr} 

\usepackage{soul}
\usepackage[dvipsnames]{xcolor}
\usepackage{colortbl}

\newcommand{\TODO}[1]{\textbf{\color{red}[TODO: #1]}}
\newcommand{\todo}[1]{{\color{RedOrange}\textbf{TODO}: #1}}
\newcommand{\myparagraph}[1]{\vspace{0.5em}\noindent\textbf{#1}}

\newcommand{\Tref}[1]{Table~\ref{#1}}

\newcommand{\Fref}[1]{Figure~\ref{#1}}
\newcommand{\sref}[1]{Sec.~\ref{#1}}

\def\eg{\emph{e.g}} 
\def\ie{\emph{i.e}}

\newcommand{\norm}[1]{\left\lVert#1\right\rVert}

\newcommand{\xT}{{\vx_T}}
\newcommand{\xt}{{\vx_t}}
\newcommand{\xtone}{{\vx_{t-1}}}

\newcommand{\x}{\mathbf{x}}

\newcommand{\I}{\mathbf{I}}

\definecolor{SceneBlue}{rgb}{0.7031,    0.7812,    0.8632}
\definecolor{SceneRed}{rgb}{0.8867,    0.6171,    0.5781}
\definecolor{SceneOrange}{rgb}{0.9296,    0.8125,    0.5859}
\definecolor{ScenePurple}{rgb}{0.7265,    0.6289,    0.7773}
\definecolor{SceneGreen}{rgb}{0.6562,    0.7656,    0.5937}

\newcommand{\shortname}{HOI-Diff}

\usepackage{amsmath,amsfonts,bm}

\def\eqref#1{equation~\ref{#1}}

\def\1{\bm{1}}

\def\rmW{{\mathbf{W}}}

\def\vmu{{\bm{\mu}}}

\def\vc{{\bm{c}}}
\def\vd{{\bm{d}}}

\def\vf{{\bm{f}}}

\def\vp{{\bm{p}}}

\def\vx{{\bm{x}}}
\def\vy{{\bm{y}}}

\DeclareMathAlphabet{\mathsfit}{\encodingdefault}{\sfdefault}{m}{sl}
\SetMathAlphabet{\mathsfit}{bold}{\encodingdefault}{\sfdefault}{bx}{n}

\newcommand{\E}{\mathbb{E}}

\newcommand{\R}{\mathbb{R}}

\usepackage{graphicx}
\usepackage{placeins}

\usepackage[utf8]{inputenc} 
\usepackage[T1]{fontenc}    
\usepackage{url}            
\usepackage{booktabs}       
\usepackage{amsfonts}       
\usepackage{nicefrac}       
\usepackage{microtype}      
\PassOptionsToPackage{prologue,dvipsnames}{xcolor}

\definecolor{cvprblue}{rgb}{0.21,0.49,0.74}
\usepackage[pagebackref,breaklinks,colorlinks,allcolors=cvprblue]{hyperref}

\usepackage{floatrow}
\usepackage{multirow}
\usepackage{algorithm}
\usepackage{algpseudocode}
\usepackage{wrapfig}
\usepackage[title]{appendix}%
\newcommand\extrafootertext[1]{%
    \bgroup
    \renewcommand\thefootnote{\fnsymbol{footnote}}%
    \renewcommand\thempfootnote{\fnsymbol{mpfootnote}}%
    \footnotetext[0]{#1}%
    \egroup
}

\newcommand{\mypar}[1]{\vspace{1mm}\noindent\textbf{#1}}

\newcommand{\new}[1]{{\color{black} #1}}
\newcommand\blfootnote[1]{%
  \begingroup
  \renewcommand\thefootnote{}\footnote{#1}%
  \addtocounter{footnote}{-1}%
  \endgroup
}

\newif\ifdrafting
\draftingtrue 
\ifdrafting
    \newcommand{\yx}[1]{\textcolor{orange}{YX: #1}}
    \newcommand{\hj}[1]{\textcolor{red}{HJ: #1}} 
    \newcommand{\ds}[1]{{\color{blue}[DS: #1]}}
    \newcommand{\VJ}[1]{{\color{magenta}[VJ: #1]}}
    \newcommand{\xp}[1]{{\color{purple}[XP: #1]}}
    
\else
	\newcommand{\yx} [1] {}
	\newcommand{\hj} [1] {}
	\newcommand{\ds} [1] {}
        \newcommand{\VJ} [1] {}
        \newcommand{\xp} [1] {}
        \renewcommand{\st} [1] {}
        \renewcommand{\todo} [1] {}
        \renewcommand{\TODO} [1] {}
\fi

\def\paperID{*****} 
\def\confName{CVPR}
\def\confYear{2025}

\title{\shortname: Text-Driven Synthesis of 3D Human-Object Interactions using Diffusion Models}

\author{
  Xiaogang Peng$^{1*}$, 
  Yiming Xie$^{1*}$,
  Zizhao Wu$^{2}$,
  Varun Jampani$^{3}$, 
  Deqing Sun$^{4}$,
  Huaizu Jiang$^{1}$
  \\
  $^{1}$Northeastern University \quad
  $^{2}$Hangzhou Dianzi University\\
  $^{3}$Stability AI \quad
  $^{4}$Google Research\\
  \texttt{\url{https://neu-vi.github.io/HOI-Diff/}}
}

\begin{document}

\twocolumn[{%
\renewcommand\twocolumn[1][]{#1}%
    \maketitle
    \centering
    \includegraphics[width=0.85\linewidth]{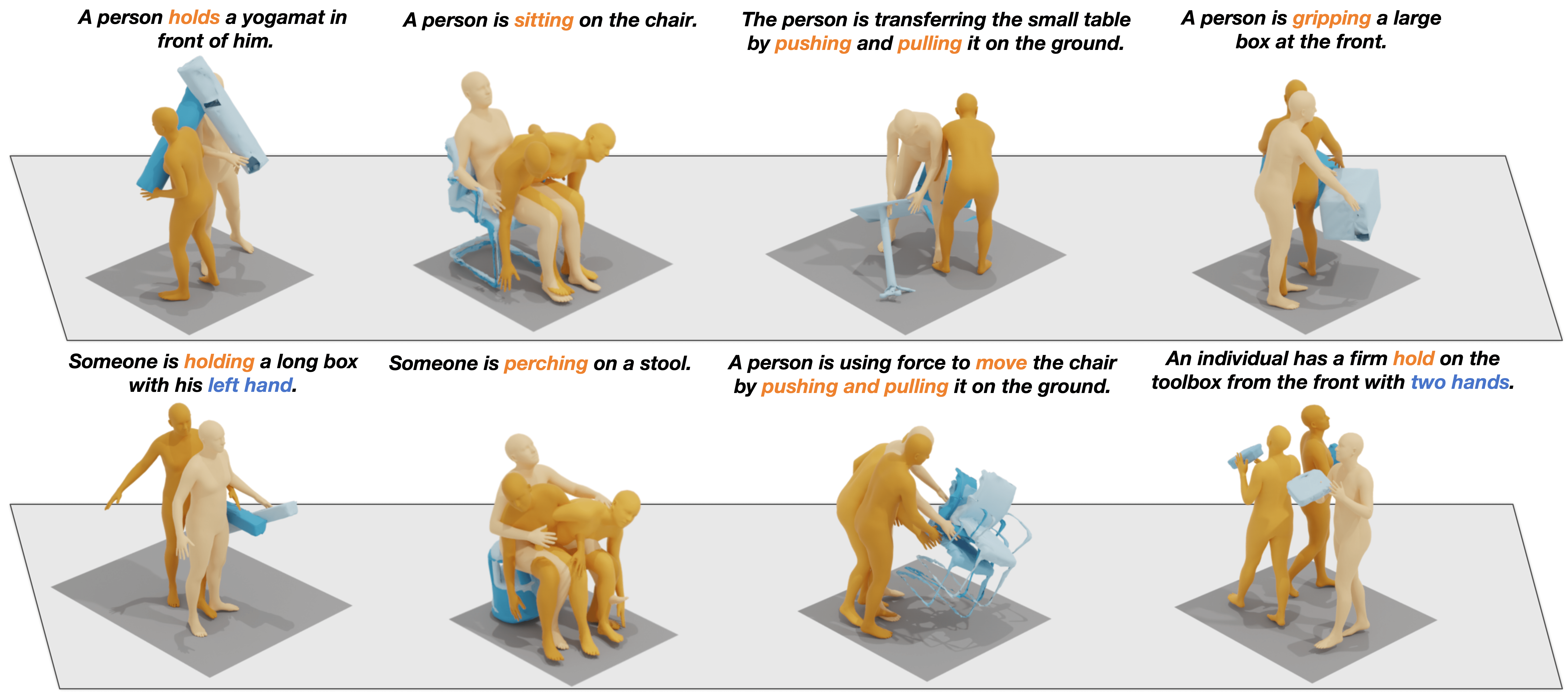}
    \captionof{figure}{
            \textbf{\shortname} \textbf{can generate realistic motions for 3D human-object interactions given a text prompt and object geometry.}
            Please see the supplementary material for video results.
            \emph{Darker color indicates later frames in the sequence. Best viewed in color.} 
            }
        \vspace{15pt}
    \label{fig:teaser}
}]

\blfootnote{$^*$The first two authors contributed equally.}

\begin{abstract}
We address the problem of generating realistic 3D human-object interactions (HOIs) driven by textual prompts. 
To this end, we take a modular design and decompose the complex task into simpler sub-tasks. 
We first develop a dual-branch diffusion model (DBDM) 
to generate both human and object motions conditioned on the input text, and encourage coherent motions by a cross-attention communication module between the human and object motion generation branches. 
We also develop an affordance prediction diffusion model (APDM) to predict the contacting area between the human and object during the interactions driven by the textual prompt.
The APDM is independent of the results by the DBDM and thus can correct potential errors by the latter. 
Moreover, it stochastically generates the contacting points to diversify the generated motions. 
Finally, we incorporate the estimated contacting points into the classifier-guidance to achieve accurate and close contact between humans and objects.
To train and evaluate our approach, we annotate the BEHAVE dataset with text descriptions.
Experimental results on BEHAVE and OMOMO demonstrate that our approach produces realistic HOIs with various interactions and different types of objects.

\end{abstract}    
\vspace{-3mm}
\section{Introduction}
\vspace{-2mm}
\label{sec:intro}
Text-driven synthesis of 3D human-object interactions 
(HOIs) aims to generate motions for both the human and object that form coherent and semantically meaningful interactions.
It enables virtual humans to naturally interact with objects, which has a wide range of applications in AR/VR, video games, and filmmaking, etc.


The generation of natural and physically plausible 3D HOIs involves humans interacting with \textit{dynamic} objects in \textit{various} ways according to the text prompts, thereby posing several challenges.
First, 
the variability of object shapes makes it particularly challenging to generate semantically meaningful contact between the human and object to avoid floating objects. 
Second, the generated HOIs should be faithful to the input text prompts as there are many plausible interactions between human and the same object (\eg, a person carries a chair, sits on a chair, pushes or pulls a chair). 
Text-driven 3D HOI synthesis with a diverse set of interactions is not yet fully addressed.
Third, the development and evaluation of 3D HOI synthesis models requires a high-quality human motion dataset with various HOIs and textual descriptions, but existing datasets lack either diverse HOIs~\citep{Guo_2022_CVPR,kit2016,li2023controllable} or detailed textual descriptions with interacting body parts and action~\citep{bhatnagar22behave, diller2023cghoi}. It is important to note that CG-HOI~\citep{diller2023cghoi} has not made their code or annotations publicly available. In contrast, we will release both our code and annotations.

Current methods cannot fully handle all the challenges.
On one hand, recent methods~\citep{kulkarni2023nifty,jiang2022chairs,hassan_samp_2021,starke2019neural,zhang2022couch,wu2022saga,taheri2021goal,Pi_2023_ICCV} can synthesize realistic human motions for HOIs for \textit{static} objects only.
They usually synthesize the motion in the last mile of interaction, \ie, the motion between the given starting human pose and the final interaction pose, and overlook the movement of the objects when the human is interacting with them. 
On the other hand, existing methods for motion generation with dynamic objects do not adequately reflect real-world complexity.
For instance, they focus on grasping small objects~\citep{ghosh2023imos}, provide the object motion as conditioning~\citep{li2023object}, predict deterministic interactions between the human and the same object without the diversity~\citep{xu2023interdiff,razali2023action}, 
consider only a small set of interactions (\eg., sit/lift~\citep{kulkarni2023nifty}, sit/lie down~\citep{hassan_samp_2021}, sit~\citep{jiang2022chairs,zhang2022couch,Pi_2023_ICCV}, grasp~\citep{wu2022saga,taheri2021goal}), or investigate a single type of object (\eg., chair~\citep{jiang2022chairs,zhang2022couch}).

In this paper, we introduce \textbf{\shortname}~for 3D HOIs synthesis involving humans interacting with different types of objects in diverse ways, which are both physically plausible and semantically faithful to the textual prompt, as shown in \Fref{fig:teaser}.
Our key insight is to decompose 3D HOIs synthesis into three modules to reduce the complexity of this challenging task.
(a) \textbf{coarse 3D HOIs generation} that extends the human motion diffusion model~\citep{tevet2023human} to a dual-branch diffusion model (DBDM) to generate both human and object motions conditioning on the input text prompt.
To encourage coherent motions, we develop a cross-attention communication module, exchanging information between the human and object motion generation models;
(b) \textbf{affordance prediction diffusion model} (APDM) that estimates the contacting points between the human and object during the interactions driven by the textual prompt.
Our APDM does not rely on the results of the DBDM and thus can recover from its potential errors. 
Moreover, it stochastically generates the contacting points to diversity the generated motions;
and (c) \textbf{affordance-guided interaction correction} that incorporates the estimated contacting information and employs the classifier-guidance to achieve accurate and close contact between humans and objects, 
significantly alleviating the cases of floating objects.
Compared with designing a monolithic model, \shortname~disentangles motion generation for humans and objects and estimation of their contacting points, which are later integrated to form coherent and diverse HOIs, reducing the complexity and burden for each of the three modules.

For both training and evaluation purposes, we annotate each video sequence in BEHAVE dataset~\citep{bhatnagar22behave} with text descriptions, which mitigates the issue of severe data scarcity for text-driven 3D HOIs generation.
In addition, we evaluate our approach on the OMOMO dataset~\citep{li2023object}, which focuses on the manipulation of two hands.
Extensive experiments validate the effectiveness and design choices of our approach, particularly for dynamic objects, thereby enabling a set of new applications in human motion generation.

\vspace{-2mm}
\section{Related Work}
\label{sec:related}
\vspace{-2mm}
\mypar{Human Motion Generation with Diffusion Models.}
The denoising diffusion models have been widely used 2D image generations~\citep{rombach2022high,saharia2022photorealistic,ramesh2021zero} and achieved impressive results.
Recent work~\citep{zhang2022motiondiffuse,tevet2023human,chen2023executing,karunratanakul2023gmd,rempeluo2023tracepace,ahn2023can,barquero2023belfusion,chen2023humanmac,dabral2023mofusion,shafir2023human,sun2023towards,tian2023transfusion,wei2023understanding,zhang2023generating,zhang2023remodiffuse,zhang2023tedi,xie2023omnicontrol} apply the diffusion model in the task of human motion generation.
While these methods have successfully generated human motion, they usually generate isolated motions in the free space without considering the objects the human is interacting with.
Our method is primarily focused on motion generation with human-object interactions.

\vspace{-0.2em}
\mypar{Scene- and Object-Aware Human Motion Generation.}
Recent works condition motion synthesis on scene geometry~\citep{huang2023diffusion,zhao2023synthesizing,wang2022towards,wang2022humanise}.
This facilitates the understanding of human-scene interactions. 
However, the motion fidelity is compromised due to the lack of paired full scene-motion data.
Other approaches~p\cite{kulkarni2023nifty,jiang2022chairs,hassan_samp_2021,starke2019neural,zhang2022couch,Pi_2023_ICCV} instead focus on the interactions with the objects and can produce realistic motions.
However, they focus on interacting with static objects with limited interactions.
OMOMO~\citep{li2023object}
can generate full-body motion from the object motion.
The object motion is needed as input in OMOMO, whereas our method can jointly synthesize human motion and object motion.
IMoS~\citep{ghosh2023imos} synthesizes the full-body human along with the 3D object motions from textual inputs,
but it only focuses on grasping small objects with hands.
InterDiff~\citep{xu2023interdiff} predicts whole-body interactions with dynamic objects. 
Note that the interaction type is deterministic.
Different from this, we tackle the motion synthesis task, where the interaction with the same object can be controlled by the text prompt. 
Recently, there has been a surge of interest in the text-driven synthesis of 3D human-object interactions for dynamic objects, resulting in the development of concurrent works~\citep{diller2023cghoi,wang2023physhoi,li2023controllable, song2024hoianimator, xu2024interdreamer}.
CG-HOI~\citep{diller2023cghoi} and HOIAnimator~\citep{song2024hoianimator} uses SMPL parameters as the motion representation, which may result in unsmooth motion due to the potential difficulty in optimization.
Instead, we use common skeletal joints similar to most text-to-motion methods, harnessing the power of pre-trained human motion generation models. 
Chois~\citet{li2023controllable} relies on the initial state and object waypoints to generate HOIs, which reduces motion diversity for both the human and the object. InterFusion~\citep{dai2024interfusion} and F-HOI~\citep{yang2024f} generate static 3D HOIs from text description, lacking both human and object motions.

\begin{figure*}[t]
    \centering
    \includegraphics[width=.75\linewidth]{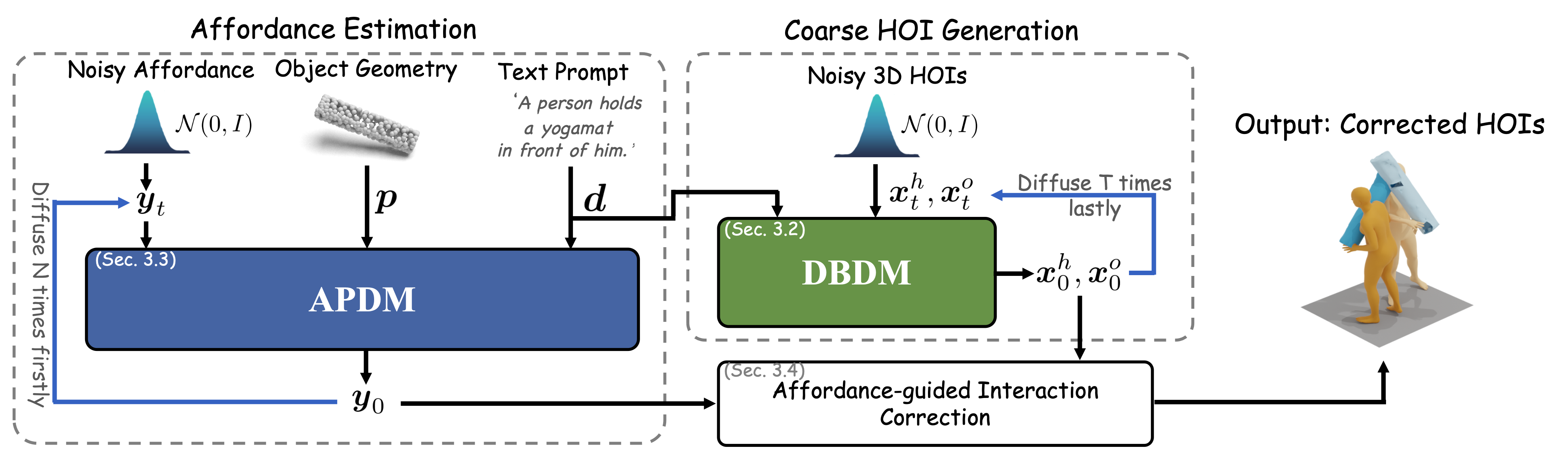}
    \vspace{-2mm}
    \caption{\textbf{Overview of HOI-Diff for 3D HOIs generation using diffusion models.}
    Our key insight is to decompose the generation task into three modules: (a) coarse 3D HOI generation using a dual-branch diffusion model (DBDM), (b) affordance prediction diffusion model (APDM) to estimate the contacting points of humans and objects, and (c) affordance-guided interaction correction, which incorporates the estimated contacting information and employs the classifier-guidance to achieve accurate and close contact between humans and objects to form coherent HOIs.
      }
    \label{fig:pipeline_overview}
\end{figure*}

\vspace{-0.2em}
\mypar{Affordance Estimation.} 
The affordance estimation on 3D point cloud is studied in~\citet{ngyen2023open,deng20213d,kokic2017affordance,iriondo2021affordance,mo2022o2o,kim2014semantic,kim2015interactive}.
Overall affordance learning is a very challenging task.
Instead of predicting the point-wise contact labels, 
we 
simplify it by directly regressing the contact points for human-object interactions, making it more tractable without significantly compromising accuracy. 

\vspace{-2mm}
\section{Method}\label{sec:method}
\vspace{-2mm}



The overview of our proposed approach are illustrated in~\Fref{fig:pipeline_overview}.
We introduce a dual-branch Human-Object Interaction Diffusion Model (DBDM), which can produce diverse yet consistent motions, capturing the intricate interplay and mutual interactions between humans and objects (Sec.~\ref{method:hoigenerate}).
To ensure physically plausible contact between humans and objects, we propose a novel affordance prediction diffusion model (APDM) (Sec.~\ref{method:affordance}), whose output will be used as classifier guidance (Sec.~\ref{method:guidance}) to correct the interactions at each diffusion step of human/object motion generation.

\vspace{-2mm}
\subsection{Background}
\label{method:background}
\vspace{-2mm}
\mypar{Motion Representations.}
We denote a 3D HOI sequence as $\vx = \{\vx^{h}, \vx^{o}\}$.
It consists of human motion sequence $\vx^{h} \in \R^{L\times D^h}$ and object motion sequence $\vx^{o} \in \R^{L \times D^o}$, where $L$ denotes the length of the sequence. 
For $\vx^h$, we adopt the redundant representation widely used in human motion generation~\citep{Guo_2022_CVPR} with $D^h = 263$, which include pelvis velocity, local joint positions, velocities and rotations of other joints in the pelvis space, and binary foot-ground contact labels. 
For the object motion sequence $\vx^o$, we assume the object geometry is given as an input, and thus we only need to estimate its 6DoF poses in the generation, \ie, $D^o = 6$.
We represent each object instance as a point cloud of 512 points $\vp\in\R^{512 \times 3}$.

\mypar{Diffusion Model for 3D HOI Generation.}
Given a prompt $\vc=(\vd, \vp)$, consisting of a textual description $\vd$ and the object instance's point cloud $\vp$, a diffusion model $p_\theta(\xtone|\xt, \vc )$\footnote{We use superscripts $h$ and $o$ to denote human and object sequence, respectively. Without a superscript, it means the 3D HOI sequence, containing both $\vx^h$ and $\vx^o$. Subscript is used for the diffusion denoising step.} learns the reverse diffusion process to generate clean data from a Gaussian noise $\xT$ with $T$ consecutive denoising steps 
\begin{equation}
    p_\theta(\xtone|\xt, \boldsymbol c) := \mathcal{N}(\xtone, \mathbf{\mu}_{\theta}(\xt, t, \boldsymbol c), (1 - \alpha_t) \I),
\label{eq:hoi_diffusion}
\end{equation}
where $t$ is the denoising step.
Following~\cite{tevet2023human}, 
our diffusion model $M_{\theta}$ with parameters $\theta$ predicts the final clean motion 
$\vx_0 = M_{\theta}(\xt, t, \vc)$. 

We sample $\x_{t-1} \sim \mathcal{N}(\vmu_t, \Sigma_t)$ and compute the mean as in~\cite{nichol2021improved}
\begin{align}
    \vmu_{t} = \frac{\sqrt{\alpha_{t-1}} \beta_t}{1-\alpha_t}\vx_0 + \frac{\sqrt{1-\beta_t}(1-\alpha_{t-1})}{1-\alpha_t}\xt,
    \label{eq:mu}
\end{align} 
where $\alpha_t = \prod_{s=1}^t (1-\beta_s)$ and $\beta_t \in(0,1)$ are the variance schedule. $\Sigma_t = \frac{1-\alpha_{t-1}}{1-\alpha_t}\beta_t$ \citep{Ho2020-ew} is a variance scheduler of choice. Similar to $\vx_t$, $\vmu_t$ consists of $\vmu^h_t$ and $\vmu^o_t$, corresponding to human and object motion, respectively.


Simply adopting the diffusion model described in Eq.(\ref{eq:hoi_diffusion}) would impose a huge burden on the model, which requires joint generation of human and object motion and more critically, enforcement of their intricate interactions to follow the input textual description.
In this paper, we propose \textbf{\shortname} for 3D HOIs generation, disentangling motion generation for humans and objects and estimation of their contacting points.
They are later integrated to form coherent and diverse HOIs, which reduces the complexity and burden for each of the three modules, leading to better generation performance as evidenced by our experiments.



\vspace{-2mm}
\subsection{Coarse 3D HOIs Generation}
\vspace{-2mm}
\label{method:hoigenerate}

\begin{figure*}[t]
    \begin{floatrow}
    \ffigbox{
        \includegraphics[width=0.7\linewidth]{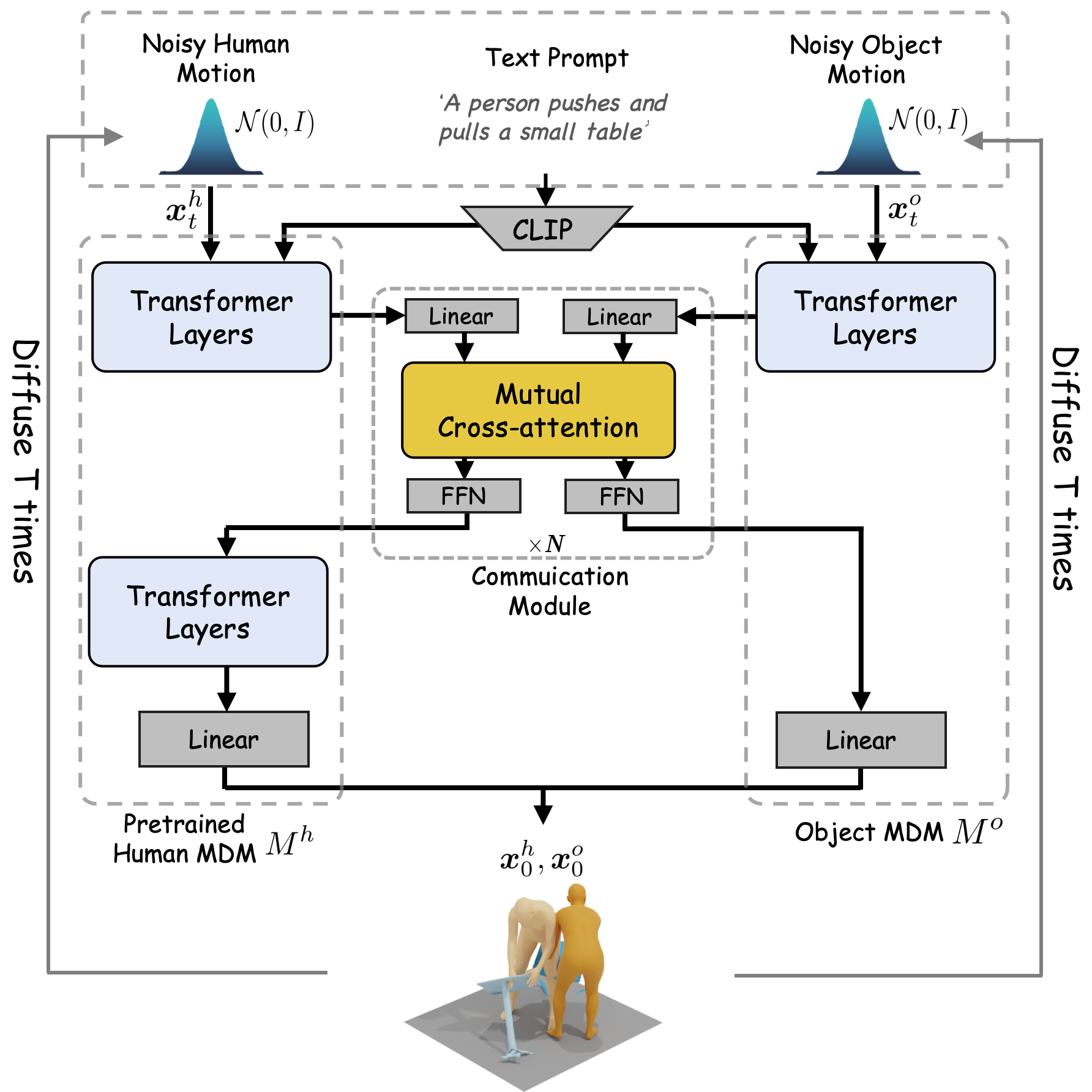}
    }{
      \caption{\textbf{Illustration of DBDM architecture for coarse 3D HOIs generation.}
        It has two branches designed for generating human and object motions individually.
        A mutual cross-attention is introduced to allow information exchange between two branches to generate coherent motions.
        The human motion model $M^{h}$ finetunes a pretrained MDM~\citep{tevet2023human}.
        }
        \label{fig:hoi_dm}
    }
    \ffigbox{
      \includegraphics[width=0.8\linewidth]{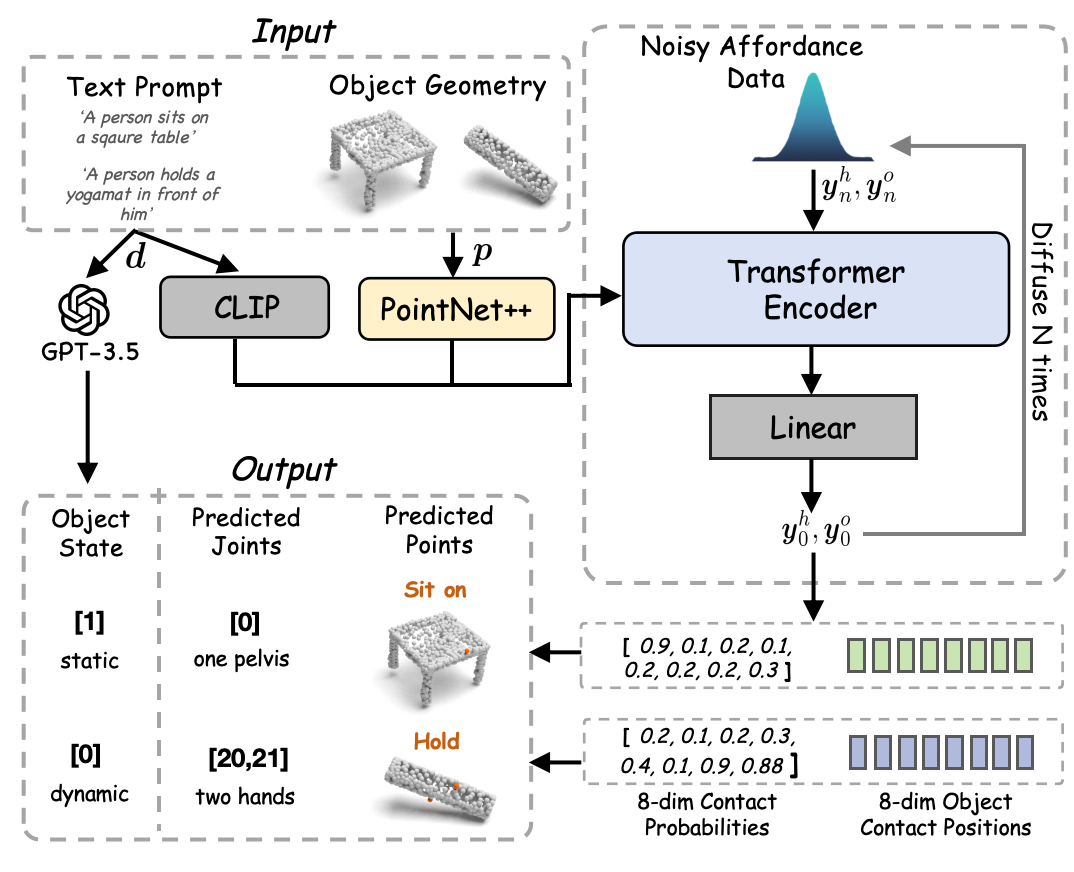}
    }{
      \caption{\textbf{Illustration of APDM architecture for affordance estimation.} Affordance information of human contact labels, object contact positions,  and binary object states are represented together as a noise variable, which is fed into the Transformer encoder to generate clean estimation. The object point cloud and textual prompt are taken as conditional input.
      }
      \label{fig:affordance_estimation}
    }
    \end{floatrow}
\end{figure*}

First, we introduce a dual-branch diffusion model (DBDM) to generate human and object motions that are roughly coherent.
As shown in ~\Fref{fig:hoi_dm}, it consists of two Transformer models~\citep{vaswani2017attention}, human motion diffusion model (MDM) $M^h$ and object MDM $M^o$, which work similar to~\cite{tevet2023human}.
Specifically, at the diffusion step $t$, they take the text description and noisy motions $\vx_t^h$ and $\vx_t^o$ as input and predict clean human and object motions $\vx_0^h$ and $\vx_0^o$,  respectively.

To enhance the learning of interactions of the human and object when generating their motion, we introduce a Communication Module ($CM$) designed for exchanging feature representations between the human MDM $M^h$ and the object MDM $M^{o}$. 
$CM$ is a Transformer block that receives the intermediate feature $\vf^h, \boldsymbol f^o$ from both $M^h$ and $M^o$. 
It then processes these inputs to generate refined updates based on the cross attention mechanism~\citep{vaswani2017attention}.
The updated feature representations
$\tilde{\vf_h}$ and $\tilde{\vf_o}$ of the human and object are then conditioned on each other, which are then fed into the subsequent layers of their respective branches to estimate clean human and object motion $\vx_0^h$ and $\vx_0^o$, respectively.
The $CM$ is inserted at the 4th transformer layer for human MDM and the last layer for object MDM, which was empirically found to work better.

Given the limited data availability for 3D HOI generation, during training, the human motion model $M^{h}$ finetunes a pretrained human MDM~\citep{tevet2023human}. 
This fine-tuning is critical to ensure the smoothness of the generated human motions. 
We ablate this design choice in~\sref{ablation}.
Object MDM is trained from scratch. We modify the input and output linear layers to take in the object motion which has a different dimension from the human motion.
More details of DBDM are in Appendix~\ref{supp:dbdm}.

\vspace{-1em}
\vspace{1mm}
\subsection{Affordance Estimation}
\vspace{-2mm}
\label{method:affordance}
Due to the complexity of the interactions between a human and object,
DBDM alone usually fails to produce physically plausible results, leading to floating objects or penetrations.
To improve the generation of intricate interactions, the problem that needs to be solved is to \textit{identify where the contacting areas are} between the human and object.
InterDiff~\citep{xu2023interdiff} defines the contacting area based on the distance measurement between the surface of human and object. 
This approach, however, heavily relies on the quality of the generated human and object motions and cannot recover from errors in the coarse 3D HOI results.
In addition, the contact area is diverse even with the same object and interaction type, \eg, ``sit'' can happen on either side of a table.
To this end, we introduce an Affordance Prediction Diffusion Model (APDM) for affordance estimation. 
As illustrated in~\Fref{fig:affordance_estimation}, the input includes a text description $\vd$ and the object point cloud $\vp$.
Our APDM doesn't rely on the results of the DBDM and thus can recover from the potential errors in DBDM. In addition, it stochastically generates the contacting points to ensure the diversity of the generated motions.


Affordance estimation in 3D point clouds itself is a notably challenging problem~\citep{ngyen2023open,deng20213d,kokic2017affordance,iriondo2021affordance,mo2022o2o,kim2014semantic,kim2015interactive}, especially in the context of 3D HOI generation involving textual prompt.
In this paper, we consider eight primary body joints -- the \texttt{pelvis, neck, feet, shoulders}, and \texttt{hands} -- as the interacting parts in HOI scenarios. 
It can effectively model common interactions such as grasping an object with both hands, sitting actions involving the pelvis and back, or lifting with a single hand. 
We use binary contact labels to determine which joints are in contact with the object. 
Subsequently, we predict eight corresponding contact points on the object surface, identified as the points closest to the selected body joints. Note that the binary contact label estimation for different body joints are independent, allowing us to handle complex HOIs.


Specifically, at each diffusion time step $n$ of APDM\footnote{We note that APDM and DBDM work independently. We thus use two symbols to denote the different diffusion time steps to avoid confusion.}, 
the noisy data consists of human contact labels representing the contact status for the eight primary body joints, denoted as $ \boldsymbol y^h_n \in \{0,1\}^{8}$, and the eight corresponding contact points on the object surface, denoted as $ \boldsymbol y_n^{o} \in \R^{8\times 3}$. The model is designed to predict both contact probabilities and contact positions. Subsequently, dynamic selection of contacting body joints is performed by considering predicted probabilities over a specific threshold $\tau$ (set to be 0.6). The corresponding contact points on the object are then determined based on the selected joints. APDM works similar to the diffusion denoising process described in Eq.(\ref{eq:hoi_diffusion}).
Besides, we utilize a large language model (ChatGPT) to determine whether the object state $\vy_0^s\in\{0,1\}$ should be set to static  ($\vy_0^s=1$) based on the textual description, which can help us better process static objects when synthesizing 3D HOIs, as discussed in the following section.
All the clean affordance data is grouped as $ \vy_0 = ( \vy^h_0,  \vy^o_0, \vy^s_0)$. More implementation details are in Appendix~\ref{supp:apdm}.

\vspace{-1mm}
\subsection{Affordance-guided Interaction Correction}
\vspace{-2mm}
\label{method:guidance}
With the estimated affordance, we can better align human and object motions to form coherent interactions.
To this end, we propose to use the classifier guidance~\citep{dhariwal2021diffusion} to achieve accurate and close contact between humans and objects, significantly alleviating the cases of floating objects.

Specifically, in a nutshell, we define an analytic function $G(\vmu^h_t, \vmu^o_t, \vy_0)$ that assesses how closely the generated human joints and object's 6DoF pose align with a desired objective.
In our case, it enforces the contact positions of human and object to be close to each other and their motions are smooth temporally. 
Based on the gradient of $G(\vmu^h_t, \vmu^o_t, \vy_0)$, we can perturb the generated human and object motion at each diffusion step $t$ as in~\cite{xie2023omnicontrol, karunratanakul2023guided},
\begin{align}
    \mathbf{\vmu}^h_t &= \mathbf{\vmu}^h_t - \tau_1\Sigma_t \nabla_{\vmu^h_t} G(\vmu^h_t, \vmu^o_t, \vy_0),\\ 
    \mathbf{\vmu}^o_t &= \mathbf{\vmu}^o_t - \tau_2\Sigma_t \nabla_{\vmu^o_t} G(\vmu^h_t, \vmu^o_t, \vy_0).
\end{align}
Here $\tau_1$ and $\tau_2$ are different strengths to control the guidance for human and object motion, respectively.
Due to the sparseness of object motion features, we assign a larger value to $\tau_2$ compared to $\tau_1$. 
This applies greater strength to perturb object motion, facilitating feasible corrections for contacting joints.
During the denoising stage, to eliminate diffusion models' bias that can suppress the guidance signal, we iteratively perturb $K$ times in the last denoising step. The details are illustrated in Algorithm~\ref{alg:guidance} of Appendix.


How can we define the objective function $G(\vmu^h_t, \vmu^o_t, \vy_0)$? We consider three terms here.
First, in the generated 3D HOIs, the human and object should be close to each other on the contacting points.
We therefore minimize the distance between human contact joints and object contact points 
\begin{align}
    G_{con} = \sum_{i\in\{1,2,...,8\}} \norm{R\big(\vmu^{h}_t(i)\big)  - V\big(\vmu^{o}_t, \vy^{o}_t(i)\big)}^2,
\end{align}
where $\vmu^{h}_t(i)$ and $\vy^{o}_t(i)$ denote the $i$-th available contacting joint indexed by $\vy^h_0$ and $i$-th object contact point, respectively. 
$R(\cdot)$ converts the human joint’s local positions to global absolute locations, and $V(\cdot)$ obtains the object's contact point sequence from the predicted mean of object pose $\vmu^{o}_t$.


Second, the generated motion of dynamic objects typically follows human movement. 
However, we observe that when the human interacts with a static object, such as sitting on a chair, the object appears slightly moved.
To address this, we immobilize the object's movement in the generated samples if the state is static ($\vy_0^s=1$), ensuring that proper contact is established between the human and the static object.  
The objective is defined as
\begin{equation}
\small{
G_{sta} =  \vy_0^s \cdot \sum_{l=1}^L \norm{\vmu_{t}^o(l) - \bar{\vmu}_t^o}^2,
}
\end{equation}
where $\vmu_{t}^o(l)$ denotes the object's 6DoF pose in the $l$-th frame. $\bar{\vmu}_t^o=\frac{1}{L}\sum_l \vmu_{t}^o(l)$, which is the average of predicted means of the object's pose.


Third, we define a smoothness term $G_{smo}(\mu)$ for the object motion to mitigate motion jittering during contact.
Due to the space limit, we explain it in Appendix~\ref{supp:interaction_correction}.

Finally, we combine all these goal functions to as the final objective 
\begin{align}
G = G_{con} + \alpha  G_{sta} + \beta  G_{smo},
\end{align}
where $\alpha=500$ and $\beta=100$ are weights for balance. 

\vspace{-2mm}
\section{Experiments}\label{sec:exp}
\vspace{-2mm}
\subsection{Setup}
\vspace{-2mm}
\myparagraph{Dataset.}
Since the data designed for studying text-driven 3D HOIs generation is severely scarce,
we manually label interaction types, interacting subjects, and contact body parts on top of the BEHAVE dataset ~\citep{bhatnagar22behave}. We then use GPT-3.5~\citep{chatgpt} to rephrase and generate three text descriptions for each HOI sequence, increasing the diversity of the data.
Specifically, BEHAVE encompasses the interactions of 8 subjects with 20 different objects. 
It provides the human SMPL-H representation~\citep{SMPL:2015}, the object mesh, as well as its 6DoF pose information in each HOI sequence. 
To ensure consistency in our approach, we follow the processing method used in HumanML3D~\citep{Guo_2022_CVPR} to extract representations for 22 body joints. 
All the models are trained to generate $L=196$ frames in our experiments.
In the end, we have 1451 3D HOI sequences along with textual descriptions to train and evaluate our proposed approach. 
We follow the official train/test split on BEHAVE.
We provide more details of the dataset and annotation process in Appendix~\ref{supp:dataset}.

In addition, we evaluate our approach on OMOMO dataset~\citep{li2023object}.
OMOMO focuses on full-body manipulation with hands. It consists of human-object interaction motion for 15 objects in daily life, with a total duration of approximately 10 hours.
It provides text descriptions for each interaction motion.
We utilize their object split strategy for both training and evaluation, ensuring the objects between the training and testing sets are different. Additionally, we preprocess human and object motion, similar to our way for the BEHAVE dataset.
More details are in Appendix~\ref{supp:omomo}.

\vspace{-0.5em}
\myparagraph{Evaluation metrics.}
We first assess different models for human motion generation using standard metrics as introduced by~\citep{Guo_2022_CVPR}, namely \textit{Fréchet Inception Distance (FID)},  \textit{R-Precision}, and  \textit{Diversity}. 
\textit{FID} quantifies the discrepancy between the distributions of actual and generated motions via a pretrained motion encoder.  \textit{R-Precision} gauges the relevance between generated motions and their corresponding text prompts.  \textit{Diversity} evaluates the range of variation in the generated motions. Additionally, we compute the  \textit{Foot Skating Ratio} to measure the proportion of frames exhibiting foot skid over a threshold (2.5 cm) during ground contact (foot height $<$ 5 cm).

To evaluate the effectiveness of HOIs generation, we report the \textit{Contact Distance} metric, which quantitatively measures the proximity between the ground-truth human contact joints and the object contact points. 
Ideally, we should develop similar metrics, \eg, \textit{FID}, to evaluate the \emph{stochastic} HOI generation. 
However, due to the limited data available in BEHAVE~\citep{bhatnagar22behave}, training a motion encoder would produce biased evaluation results.
To mitigate this issue, we resort to user studies to quantify the effectiveness of different models. Details will be introduced later.



\vspace{-2.5mm}
\subsection{Comparisons with Existing Methods}
\vspace{-2mm}
\noindent\textbf{Baselines.} Our work introduces a novel 3D HOIs generation task not addressed by existing text-to-motion methods, which focus exclusively on human motion generation without accounting for human-object interactions.
To compare with existing works, we mainly focus on evaluating human motion generation. 
We then design different variants of our models for comparing 3D HOIs generation.
Specifically, we adopt the prominent text-to-motion methods MDM~\citep{tevet2023human} and PriorMDM*~\citep{shafir2023human} with the following settings. 
(a) MDM$^{\dag}$: In this setup, we finetune the original MDM model~\citep{tevet2023human} on the BEHAVE dataset~\citep{bhatnagar22behave} without object motion.
(b) MDM*: This variant involves adapting the input and output layers' dimensions of the MDM model~\citep{tevet2023human} to accommodate the input of 3D HOI sequences. This adjustment allows for the simultaneous learning of both human and object motions within a singular, integrated model. 
(c) PriorMDM*~\citep{shafir2023human}: We adapt the ComMDM architecture proposed in~\cite{shafir2023human}, originally designed for two-person motion generation, to suit our needs for HOIs synthesis by modifying one of its two branches for object motion generation.
(d) InterDiff~\citep{xu2023interdiff}: While InterDiff is not designed for text-driven synthesis of 3D HOI, we added text conditioning to InterDiff as the baseline.
More details are in Appendix~\ref{supp:baselines}.

\begin{table*}[t]
\renewcommand{\arraystretch}{1.1}
\begin{center}
\resizebox{.9\linewidth}{!}{\begin{tabular}{|l|c|c|c|c|c|c|c|c|c|c|c|c|}
\hline
& \multicolumn{6}{c|}{BEHAVE} & \multicolumn{6}{c|}{OMOMO} \\
\hline
 Method & \multicolumn{1}{p{0.8cm}|}{\centering FID  \\ $\downarrow$} & \multicolumn{1}{p{1.9cm}|}{\centering R-precision \\ (Top-3) $\uparrow$ } & \multicolumn{1}{p{1.3cm}|}{\centering Diversity  \\ $\rightarrow$} & \multicolumn{1}{p{1.5cm}|}{\centering Contact  \\ Distance $\downarrow$} &  \multicolumn{1}{p{0.8cm}|}{\centering Pene  \\ $\downarrow$} & \multicolumn{1}{p{1.5cm}|}{\centering Foot Skate \\ Ratio $\downarrow$} & \multicolumn{1}{p{0.8cm}|}{\centering FID  \\ $\downarrow$} & \multicolumn{1}{p{1.9cm}|}{\centering R-precision \\ (Top-3) $\uparrow$ } & \multicolumn{1}{p{1.3cm}|}{\centering Diversity  \\ $\rightarrow$} & \multicolumn{1}{p{1.5cm}|}{\centering Contact  \\ Distance $\downarrow$}  &  \multicolumn{1}{p{0.8cm}|}{\centering Pene  \\ $\downarrow$} & \multicolumn{1}{p{1.5cm}|}{\centering Foot Skate \\ Ratio $\downarrow$} \\
\hline
 Real  &  0.04  & 0.86  & 12.48 & - & -&  - & 0.57 &  0.63 & 9.98  & - & - & -\\
\hline
 MDM$^{\dag}$ & 6.77     & 0.34     & 10.81   & - & - & - & 12.28 & 0.23 & 5.56 & - &- & -\\
 MDM*      & 4.25    &  0.38    &  11.23   & 0.448 & 0.52   & 0.190 & 10.37 & 0.21 & 6.04 & 0.768 & {0.41} & 0.191\\
 PriorMDM*   & 4.54 & 0.30  & 10.03  &   0.416     & 0.57 & 0.270    & 9.87 & 0.25 & 6.34 & 0.523 & {0.38} & 0.344\\ 
 InterDiff  & 8.58 & 0.26   & 10.75  & 0.506       & {0.42 }& 0.218 & 14.27 & 0.17 & 5.69 & 0.906 & {0.32} & 0.239\\
 \textbf{Ours} & \textbf{1.62}    &\textbf{0.46}    &\textbf{12.02}    & \textbf{0.347}  & {0.51} & \textbf{0.182}  & \textbf{8.76} & \textbf{0.31} & \textbf{8.13} &\textbf{ 0.326}  & {0.39} & \textbf{0.141} \\
\hline
\end{tabular}}
\vspace{-6mm}
\caption{\textbf{Quantitative results on the BEHAVE and OMOMO dataset.} We compare our method with baselines adapted from existing models. MDM$^{\dag}$: fine-tune the original MDM~\citep{tevet2023human} on the BEHAVE dataset without object motion.
MDM*: adapting the input and output layers' dimensions of the MDM to accommodate both human and object motions. 
PriorMDM*: We adapt the ComMDM architecture proposed in~\citet{shafir2023human}. InterDiff: We add a CLIP encoder in~\citet{xu2023interdiff} to support our task. 
The right arrow $\rightarrow$ means closer to real data is better.}
\label{table:quant}
\end{center}
\end{table*}

\begin{figure*}[t]
    \centering
    \includegraphics[width=0.85\linewidth]{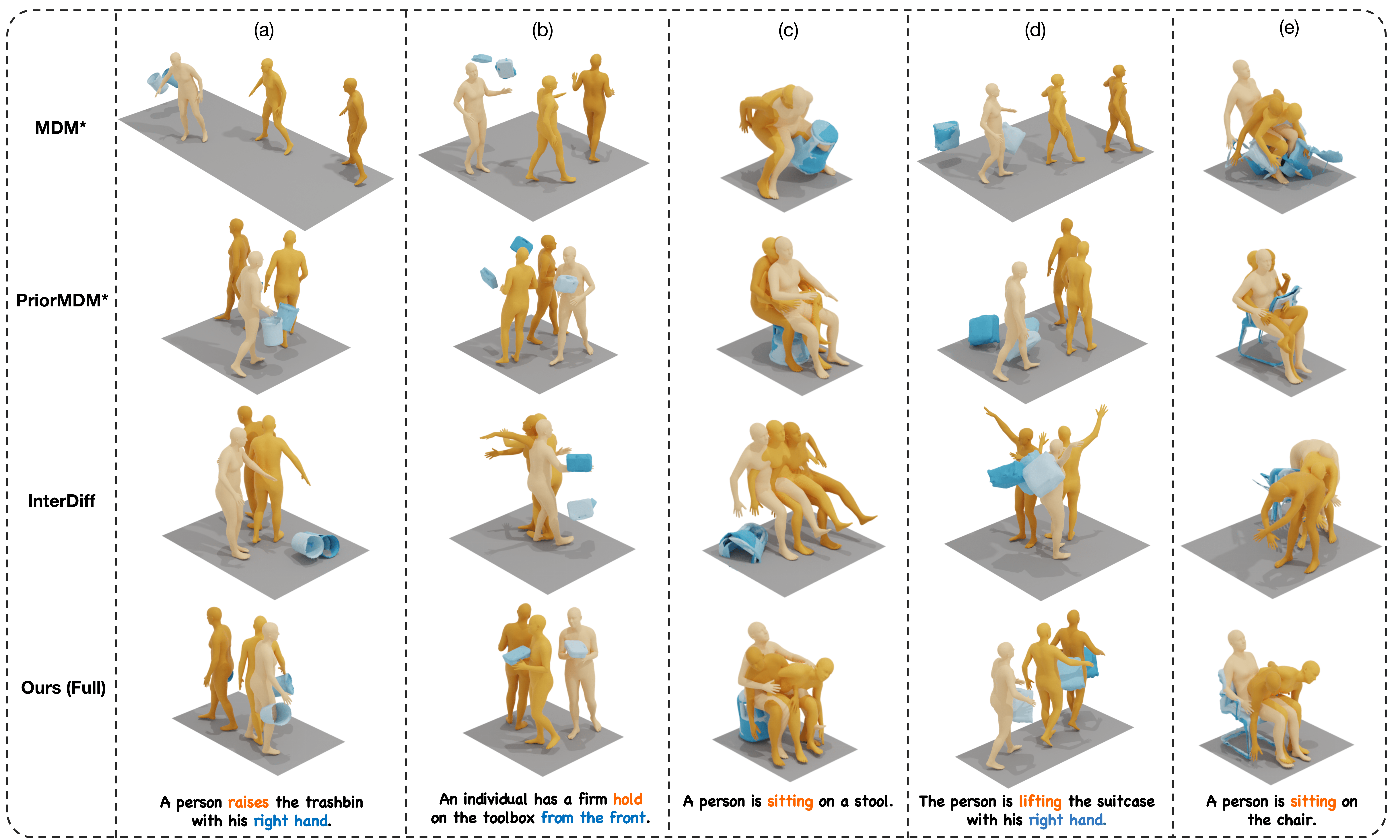}
    \caption{\textbf{Qualitative comparisons of our approach and baselines on BEHAVE dataset.} The bottom row, showcasing our method, demonstrates the generation of realistic 3D HOIs with plausible contacts, particularly evident in columns 2 and 4. This contrasts with the baselines, which fail to achieve a similar level of realism and contact plausibility in the interactions.
    As an additional visual aid, the mesh color gradually darkens over time to represent progression. (Best viewed in color.)
    }
    \label{fig:vis}
    \vspace{-2mm}
\end{figure*}

\vspace{-0.2em}
\noindent\textbf{Quantitative Results.} \Tref{table:quant}-left reports the quantitative results on BEHAVE dataset~\citep{bhatnagar22behave}. 
Compared with the baseline methods, our full method achieves the best performance. 
Specifically, it achieves state-of-the-art results in both \textit{FID}, \textit{R-precision}, and \textit{Diversity}, underscoring its ability to generate high-quality human motions in the context of coherently interacting with objects. 
The best \textit{Contact Distance} also suggests that our approach can generate physically plausible HOIs, capturing the intricate interplay interactions between humans and objects.
\Tref{table:quant}-right presents the quantitative results on the OMOMO dataset. We used the train/test split of the OMOMO dataset to evaluate the model's inference capacity on unseen objects, including the small table, white chair, suitcase, and tripod. Our method consistently outperforms other baselines by a considerable margin across all metrics. Notably, due to the distinctiveness of objects in the training and testing sets, the results indicate the effectiveness of our approach in \textit{generalizing to unseen objects}, proving superior performance compared to other models. We also provide user study results, please refer to Appendix~\ref{supp:user_study} for details.

\vspace{-0.2em}
\noindent\textbf{Qualitative Results.}
We showcase qualitative comparisons, rendered with SMPL~\citep{SMPL:2015} shapes, between our approach and the baseline methods in~\Fref{fig:vis}. It is observed that the generated HOI motion by other baselines lacks smoothness and realism, where the object may float in the air (\eg, the toolbox in~\Fref{fig:vis} (b)). 
Furthermore, these baseline methods struggle to accurately capture the spatial relationships between humans and objects (\eg, the chair in~\Fref{fig:vis} (e)). 
In stark contrast, our approach excels in creating visually appealing and realistic HOIs. Notably, it adeptly reflects the intricate details outlined in text descriptions, capturing both the nature of the interactive actions and the specific body parts involved (\eg, raising the trash bin with the right hand in~\Fref{fig:vis} (a)). \new{For the same object, our method can generate diverse HOIs using different body parts and contact points, as shown in~\Fref{fig:different_contact} in Appendix.} 

\begin{table*}[t]
\renewcommand{\arraystretch}{1.0}
\vspace{-2mm}
\begin{center}
\resizebox{.9\linewidth}{!}{\begin{tabular}{|l|c|c|c|c|c|c|c|c|c|c|}
\hline
& \multicolumn{5}{c|}{BEHAVE} & \multicolumn{5}{c|}{OMOMO} \\
\hline
 Variants & \multicolumn{1}{p{0.8cm}|}{\centering FID  \\ $\downarrow$} & \multicolumn{1}{p{1.9cm}|}{\centering R-precision \\ (Top-3) $\uparrow$ } & \multicolumn{1}{p{1.3cm}|}{\centering Diversity  \\ $\rightarrow$} & \multicolumn{1}{p{1.5cm}|}{\centering Contact  \\ Distance $\downarrow$} & \multicolumn{1}{p{1.5cm}|}{\centering Foot Skate \\ Ratio $\downarrow$} & \multicolumn{1}{p{0.8cm}|}{\centering FID  \\ $\downarrow$} & \multicolumn{1}{p{1.9cm}|}{\centering R-precision \\ (Top-3) $\uparrow$ } & \multicolumn{1}{p{1.3cm}|}{\centering Diversity  \\ $\rightarrow$} & \multicolumn{1}{p{1.5cm}|}{\centering Contact  \\ Distance $\downarrow$} & \multicolumn{1}{p{1.5cm}|}{\centering Foot Skate \\ Ratio $\downarrow$} \\
\hline
 Real  &  0.04  & 0.86  & 12.48 & - & - & 0.57 &  0.63 & 9.98  & - & -\\
\multicolumn{11}{|c|}{\cellcolor{SceneBlue}\textbf{\textit{w/o Interaction Correction}}} \\
{Ours w/o CM}            & 3.11            & 0.36             & 10.54             & 0.524           & 0.265          & 11.57           & 0.27              & 7.92           & 0.588          & 0.231\\
{Ours w/o pretrain}      & 2.98            & 0.39             & 11.21             & 0.402           & \textbf{0.158} & 10.38           & 0.29              & 7.82           & 0.412          & 0.167\\
{Ours$^{global}$}        & 15.37           & 0.28             & 10.85             & 0.375           & 0.274          & 20.22           & 0.21              & 8.02           & 0.366          & 0.348\\
{Ours }                  & 2.10            & 0.38             & 11.26             & 0.415           & 0.205          & 9.12            & 0.29              & 7.97           & 0.397          & 0.193\\
\multicolumn{11}{|c|}{\cellcolor{SceneBlue}\textbf{\textit{w/ Interaction Correction}}} \\ 
{Ours w/o ${M}^o$} \& CM & 3.93            & 0.32             & 11.43             & 0.365           & 0.310          & 11.03           & 0.28              & 7.98           & 0.536          & 0.331\\
{Ours $^{joint}$}        & 4.37            & 0.31             & 11.25             & 0.421           & 0.342          & 11.52           & 0.27              & 7.92           & 0.547          & 0.325\\
{Ours w/o $G_{con}$}     & 2.02            & 0.37             & 11.97             & 0.417           & 0.196          & 9.23            & 0.28              & 8.03           & 0.332          & 0.144\\
{Ours w/o  $G_{sta}$}    & 1.81            & 0.39             & 11.54             & 0.367           & 0.181          & 9.11            & 0.30              & 8.10           & 0.340          & 0.142\\
{Ours w/o  $G_{smo}$}    & 1.83            & 0.41             & 11.67             & 0.370           & 0.182          & 8.98            & 0.29              & 8.06           & 0.345          & 0.142\\
{Ours (Full) }           &\textbf{1.62}    & \textbf{0.46}    & \textbf{12.02}    & \textbf{0.347}  & 0.182          & \textbf{8.76 }  & \textbf{0.31}     & \textbf{8.14}  & \textbf{0.326} & \textbf{0.141}\\
\hline
\end{tabular}}
\vspace{-6mm}
\caption{
\textbf{Ablation studies of our model’s variants on the BEHAVE and OMOMO datasets.}
The right arrow → means closer to real data is better. 
\textit{w/o CM}: we remove the Communication Module (CM) in the DBDM model.
\textit{w/o pretrain}: we train human MDM from scratch on BEAHVE dataset.
\textit{$global$}: we adopt the global human pose representation proposed by ~\citet{liang2024intergen} for both the pretraining of human MDM and the finetuning of DBDM. 
\textit{w/o M$^o$ \& CM}: We exclusively finetune the human MDM, while randomly initializing the object motion. Interaction correction is then applied to optimize contact between the human and object. 
\textit{$joint$}: We train a single diffusion model that jointly generate human motion, object motion, and affordance.
\textit{\textit{w/o} $G_{con}$/$G_{sta}$/$G_{smo}$}: without contacting/static/smoothness goal function in interaction correction.
}
\label{table:ablation}
\end{center}
\end{table*}

\subsection{Ablation Studies}
\label{ablation}
\vspace{-2mm}

We conduct extensive ablation studies in \Tref{table:ablation} and \Fref{fig:ablation_vis} in Appendix to validate the effectiveness of different components. We summarize key findings below.


\new{
\vspace{-2mm}
\myparagraph{Object MDM is helpful.}
In \Tref{table:ablation}, we compare \textit{Ours w/o $M^o$ \& CM} and \textit{ours (Full)} to demonstrate the importance of the Object MDM.
In \textit{Ours w/o $M^o$ \& CM}, we exclusively finetune the human MDM, while randomly initializing the object motion. 
The Communication Module (CM) is also ignored due to the removed object MDM.
Interaction correction is then applied to optimize contact between the human and object.
The interaction correction with random initial object motion produces worse results, demonstrating the importance of initial object motion from Object MDM.

\vspace{-2mm}
\myparagraph{DBDM with Communication Module ($CM$) is critical.}
In \Tref{table:ablation}, we compare \textit{Ours w/o CM} and \textit{ours} to demonstrate the effectiveness of the Communication Module.
When eliminating $CM$, the results drop substantially across all metrics, with a particularly significant decrease in \textit{Contact Distance}.
The visual results (w/o $CM$) in \Fref{fig:ablation_vis} of Appendix further validate this point.

\vspace{-2mm}
\myparagraph{Leveraging the pre-trained Human motion prior can generate better human motions.} 
We aim to utilize the strong motion prior from the pre-trained human motion model to enhance motion realism. 
\Tref{table:ablation} (\textit{Ours w/o pretrain}) reports the results of training human MDM from scratch, without resuming the weights from the pre-trained MDM~\citep{tevet2023human}.
Comparing \textit{Ours w/o pretrain} and \textit{Ours} demonstrates the effectiveness of leveraging the pre-trained MDM.

\vspace{-2mm}
\myparagraph{Interaction Correction makes better HOIs generation.}
In \Tref{table:ablation}, we compare our full method (\textit{Ours (full)}) to a variant without interaction correction (\textit{Ours}) to demonstrate the effectiveness of interaction correction. The model with interaction correction consistently outperforms the variant across all control accuracy metrics.
As shown qualitatively in \Fref{fig:ablation_vis} of Appendix, our full method produces more realistic HOIs with better contact compared to the model without interaction correction. 
Furthermore, all sub-functions in Interaction Correction contribute to the realistic HOI generation, 
as demonstrated in \textit{Ours w/o} $G_{con}$, \textit{w/o} $G_{sta}$, \textit{w/o} $G_{smo}$ of \Tref{table:ablation}.

\vspace{-2mm}
\myparagraph{Why Human MDM and Object MDM are needed separately?}
We can ablate this by comparing \Tref{table:quant} (\textit{MDM*}) and \Tref{table:ablation} (\textit{Ours (w/o Interaction Correction)}. 
In \textit{MDM*} we jointly learn both human and object motion with a diffusion model. 
Our superior results demonstrate that separately modeling human motion and object motion with a communication module can achieve better results. 
A key advantage is that the human motion diffusion model (MDM) can fine-tune a pre-trained MDM~\citep{tevet2023human}, leveraging the extensive prior knowledge from the large-scale HumanML3D dataset. In contrast, jointly predicting human and object motion with a single transformer requires training from scratch (due to the change of the model architecture) on the much smaller BEHAVE dataset, which results in poorer 
results.

\begin{table}[h]
\renewcommand{\arraystretch}{1.1}
\vspace{-2mm}
\centering
\resizebox{0.6\linewidth}{!}{
\begin{tabular}{l|c|c}
    \toprule
    & AP (\%) $\uparrow$ & L2 Dist $\downarrow$\\ 
     \midrule
    Ours $^{joint}$ & 53.67  & 0.384\\
    Ours $^{APDM}$ & \textbf{78.54} & \textbf{0.272}\\
    \bottomrule
\end{tabular}
}
\vspace{-2mm}
\caption{APDM evaluation. The reported metrics include Average Precision (AP) for predicted human contact probabilities and L2 Distance (Dist) error for predicted object contact points.}
\label{table:APDM}
\end{table}

\vspace{-3mm}
\myparagraph{Why not jointly generate motion and affordance with one unified model?}
We attempt to generate human motion, object motion, and affordance jointly within the same model, as indicated in the \Tref{table:ablation} (\textit{Ours}$^{joint}$).
Our joint prediction concatenates affordance data with motion data along the channel dimension and adjusts the input and output dimensions of MDM to generate motions and affordance simultaneously.
Comparing \Tref{table:ablation} \textit{Ours}$^{joint}$ and \textit{Ours (full)}  shows that our modular design significantly improves human motion quality, as evidenced by metrics such as FID, R-Precision, and Foot Skate Ratio, as well as the interaction quality measured by Contact Distance.
\Tref{table:APDM} further validates that our modular design achieves more accurate affordance estimation, measured by AP and L2 Distance. 
The improvement is attributed to the fact that affordance learning is highly dependent on the geometry of 3D data and text semantics, rather than human and object motions. Therefore, disentangling these elements enhances their respective performances.


}

\vspace{-2mm}
\section{Conclusion}
\vspace{-2mm}
In summary, we presented a novel approach \shortname~to generate realistic 3D HOIs driven by textual prompts. 
By employing a modular design, we effectively decompose the complex task of HOI synthesis into simpler sub-tasks, enhancing the coherence and realism of the generated motions. Our \shortname~model successfully generates coarse dynamic human and object motions, while the affordance prediction diffusion model adds precision in predicting contact areas. The integration of estimated affordance data into classifier-guidance further ensures accurate human-object interactions.  The promising experimental results on our annotated BEHAVE dataset demonstrate the efficacy of our approach in producing diverse and realistic HOIs.

\vskip4pt \noindent{\bf Acknowledgement.~}
Yiming Xie was supported by the Apple Scholars in AI/ML PhD fellowship.


{
    \small
    \bibliographystyle{ieeenat_fullname}
    \bibliography{main}

\begin{thebibliography}{69}
\providecommand{\natexlab}[1]{#1}
\providecommand{\url}[1]{\texttt{#1}}
\expandafter\ifx\csname urlstyle\endcsname\relax
  \providecommand{\doi}[1]{doi: #1}\else
  \providecommand{\doi}{doi: \begingroup \urlstyle{rm}\Url}\fi

\bibitem[Ahn et~al.(2023)Ahn, Mascaro, and Lee]{ahn2023can}
Hyemin Ahn, Esteve~Valls Mascaro, and Dongheui Lee.
\newblock Can we use diffusion probabilistic models for 3d motion prediction?
\newblock \emph{arXiv}, 2023.

\bibitem[Azadi et~al.(2023)Azadi, Shah, Hayes, Parikh, and Gupta]{azadi2023make}
Samaneh Azadi, Akbar Shah, Thomas Hayes, Devi Parikh, and Sonal Gupta.
\newblock Make-an-animation: Large-scale text-conditional 3d human motion generation.
\newblock In \emph{Proceedings of the IEEE/CVF International Conference on Computer Vision}, pages 15039--15048, 2023.

\bibitem[Barquero et~al.(2023)Barquero, Escalera, and Palmero]{barquero2023belfusion}
German Barquero, Sergio Escalera, and Cristina Palmero.
\newblock Belfusion: Latent diffusion for behavior-driven human motion prediction.
\newblock In \emph{ICCV}, 2023.

\bibitem[Bhatnagar et~al.(2022)Bhatnagar, Xie, Petrov, Sminchisescu, Theobalt, and Pons-Moll]{bhatnagar22behave}
Bharat~Lal Bhatnagar, Xianghui Xie, Ilya Petrov, Cristian Sminchisescu, Christian Theobalt, and Gerard Pons-Moll.
\newblock Behave: Dataset and method for tracking human object interactions.
\newblock In \emph{CVPR}, 2022.

\bibitem[Chen et~al.(2023{\natexlab{a}})Chen, Zhang, Li, Pang, Xia, and Liu]{chen2023humanmac}
Ling-Hao Chen, Jiawei Zhang, Yewen Li, Yiren Pang, Xiaobo Xia, and Tongliang Liu.
\newblock Humanmac: Masked motion completion for human motion prediction.
\newblock \emph{arXiv}, 2023{\natexlab{a}}.

\bibitem[Chen et~al.(2023{\natexlab{b}})Chen, Jiang, Liu, Huang, Fu, Chen, and Yu]{chen2023executing}
Xin Chen, Biao Jiang, Wen Liu, Zilong Huang, Bin Fu, Tao Chen, and Gang Yu.
\newblock Executing your commands via motion diffusion in latent space.
\newblock In \emph{CVPR}, 2023{\natexlab{b}}.

\bibitem[Dabral et~al.(2023)Dabral, Mughal, Golyanik, and Theobalt]{dabral2023mofusion}
Rishabh Dabral, Muhammad~Hamza Mughal, Vladislav Golyanik, and Christian Theobalt.
\newblock Mofusion: A framework for denoising-diffusion-based motion synthesis.
\newblock In \emph{CVPR}, 2023.

\bibitem[Dai et~al.(2024)Dai, Li, Sun, Huang, Ma, Huang, Xu, and Hu]{dai2024interfusion}
Sisi Dai, Wenhao Li, Haowen Sun, Haibin Huang, Chongyang Ma, Hui Huang, Kai Xu, and Ruizhen Hu.
\newblock Interfusion: Text-driven generation of 3d human-object interaction.
\newblock \emph{arXiv preprint arXiv:2403.15612}, 2024.

\bibitem[Deng et~al.(2021)Deng, Xu, Wu, Chen, and Jia]{deng20213d}
Shengheng Deng, Xun Xu, Chaozheng Wu, Ke Chen, and Kui Jia.
\newblock 3d affordancenet: A benchmark for visual object affordance understanding.
\newblock In \emph{CVPR}, 2021.

\bibitem[Dhariwal and Nichol(2021)]{dhariwal2021diffusion}
Prafulla Dhariwal and Alexander Nichol.
\newblock Diffusion models beat gans on image synthesis.
\newblock In \emph{NeurIPS}, 2021.

\bibitem[Diller and Dai(2024)]{diller2023cghoi}
Christian Diller and Angela Dai.
\newblock Cg-hoi: Contact-guided 3d human-object interaction generation.
\newblock 2024.

\bibitem[Ghosh et~al.(2023)Ghosh, Dabral, Golyanik, Theobalt, and Slusallek]{ghosh2023imos}
Anindita Ghosh, Rishabh Dabral, Vladislav Golyanik, Christian Theobalt, and Philipp Slusallek.
\newblock Imos: Intent-driven full-body motion synthesis for human-object interactions.
\newblock In \emph{CGF}, 2023.

\bibitem[Guo et~al.(2022)Guo, Zou, Zuo, Wang, Ji, Li, and Cheng]{Guo_2022_CVPR}
Chuan Guo, Shihao Zou, Xinxin Zuo, Sen Wang, Wei Ji, Xingyu Li, and Li Cheng.
\newblock Generating diverse and natural 3d human motions from text.
\newblock In \emph{CVPR}, 2022.

\bibitem[Hassan et~al.(2021)Hassan, Ceylan, Villegas, Saito, Yang, Zhou, and Black]{hassan_samp_2021}
Mohamed Hassan, Duygu Ceylan, Ruben Villegas, Jun Saito, Jimei Yang, Yi Zhou, and Michael Black.
\newblock Stochastic scene-aware motion prediction.
\newblock In \emph{ICCV}, 2021.

\bibitem[Hendrycks and Gimpel(2016)]{hendrycks2016gaussian}
Dan Hendrycks and Kevin Gimpel.
\newblock Gaussian error linear units (gelus).
\newblock \emph{arXiv}, 2016.

\bibitem[Ho et~al.(2020)Ho, Jain, and Abbeel]{Ho2020-ew}
Jonathan Ho, Ajay Jain, and Pieter Abbeel.
\newblock Denoising diffusion probabilistic models.
\newblock 2020.

\bibitem[Huang et~al.(2023)Huang, Wang, Li, Jia, Liu, Zhu, Liang, and Zhu]{huang2023diffusion}
Siyuan Huang, Zan Wang, Puhao Li, Baoxiong Jia, Tengyu Liu, Yixin Zhu, Wei Liang, and Song-Chun Zhu.
\newblock Diffusion-based generation, optimization, and planning in 3d scenes.
\newblock In \emph{CVPR}, 2023.

\bibitem[Iriondo et~al.(2021)Iriondo, Lazkano, and Ansuategi]{iriondo2021affordance}
Ander Iriondo, Elena Lazkano, and Ander Ansuategi.
\newblock Affordance-based grasping point detection using graph convolutional networks for industrial bin-picking applications.
\newblock \emph{Sensors}, 2021.

\bibitem[Jiang et~al.(2022)Jiang, Liu, Cao, Cui, Chen, Wang, Zhu, and Huang]{jiang2022chairs}
Nan Jiang, Tengyu Liu, Zhexuan Cao, Jieming Cui, Yixin Chen, He Wang, Yixin Zhu, and Siyuan Huang.
\newblock Chairs: Towards full-body articulated human-object interaction.
\newblock \emph{arXiv}, 2022.

\bibitem[Karunratanakul et~al.(2023{\natexlab{a}})Karunratanakul, Preechakul, Suwajanakorn, and Tang]{karunratanakul2023gmd}
Korrawe Karunratanakul, Konpat Preechakul, Supasorn Suwajanakorn, and Siyu Tang.
\newblock Gmd: Controllable human motion synthesis via guided diffusion models.
\newblock In \emph{ICCV}, 2023{\natexlab{a}}.

\bibitem[Karunratanakul et~al.(2023{\natexlab{b}})Karunratanakul, Preechakul, Suwajanakorn, and Tang]{karunratanakul2023guided}
Korrawe Karunratanakul, Konpat Preechakul, Supasorn Suwajanakorn, and Siyu Tang.
\newblock Guided motion diffusion for controllable human motion synthesis.
\newblock In \emph{ICCV}, 2023{\natexlab{b}}.

\bibitem[Kim and Sukhatme(2014)]{kim2014semantic}
David~Inkyu Kim and Gaurav~S Sukhatme.
\newblock Semantic labeling of 3d point clouds with object affordance for robot manipulation.
\newblock In \emph{ICRA}, 2014.

\bibitem[Kim and Sukhatme(2015)]{kim2015interactive}
David~Inkyu Kim and Gaurav~S Sukhatme.
\newblock Interactive affordance map building for a robotic task.
\newblock In \emph{IROS}, 2015.

\bibitem[Kokic et~al.(2017)Kokic, Stork, Haustein, and Kragic]{kokic2017affordance}
Mia Kokic, Johannes~A Stork, Joshua~A Haustein, and Danica Kragic.
\newblock Affordance detection for task-specific grasping using deep learning.
\newblock In \emph{International Conference on Humanoid Robotics (Humanoids)}, 2017.

\bibitem[Kulkarni et~al.(2023)Kulkarni, Rempe, Genova, Kundu, Johnson, Fouhey, and Guibas]{kulkarni2023nifty}
Nilesh Kulkarni, Davis Rempe, Kyle Genova, Abhijit Kundu, Justin Johnson, David Fouhey, and Leonidas Guibas.
\newblock Nifty: Neural object interaction fields for guided human motion synthesis.
\newblock \emph{arXiv}, 2023.

\bibitem[Li et~al.(2023{\natexlab{a}})Li, Clegg, Mottaghi, Wu, Puig, and Liu]{li2023controllable}
Jiaman Li, Alexander Clegg, Roozbeh Mottaghi, Jiajun Wu, Xavier Puig, and C.~Karen Liu.
\newblock Controllable human-object interaction synthesis, 2023{\natexlab{a}}.

\bibitem[Li et~al.(2023{\natexlab{b}})Li, Wu, and Liu]{li2023object}
Jiaman Li, Jiajun Wu, and C~Karen Liu.
\newblock Object motion guided human motion synthesis.
\newblock \emph{TOG}, 2023{\natexlab{b}}.

\bibitem[Liang et~al.(2024)Liang, Zhang, Li, Yu, and Xu]{liang2024intergen}
Han Liang, Wenqian Zhang, Wenxuan Li, Jingyi Yu, and Lan Xu.
\newblock Intergen: Diffusion-based multi-human motion generation under complex interactions.
\newblock \emph{International Journal of Computer Vision}, pages 1--21, 2024.

\bibitem[Loper et~al.(2015)Loper, Mahmood, Romero, Pons-Moll, and Black]{SMPL:2015}
Matthew Loper, Naureen Mahmood, Javier Romero, Gerard Pons-Moll, and Michael~J. Black.
\newblock {SMPL}: A skinned multi-person linear model.
\newblock \emph{ACM Trans. Graphics (Proc. SIGGRAPH Asia)}, 2015.

\bibitem[Loshchilov and Hutter(2017)]{loshchilov2017decoupled}
Ilya Loshchilov and Frank Hutter.
\newblock Decoupled weight decay regularization.
\newblock In \emph{ICLR}, 2017.

\bibitem[Mo et~al.(2022)Mo, Qin, Xiang, Su, and Guibas]{mo2022o2o}
Kaichun Mo, Yuzhe Qin, Fanbo Xiang, Hao Su, and Leonidas Guibas.
\newblock O2o-afford: Annotation-free large-scale object-object affordance learning.
\newblock In \emph{CoRL}, 2022.

\bibitem[Ngyen et~al.(2023)Ngyen, Vu, Vuong, Nguyen, Vo, Le, and Nguyen]{ngyen2023open}
Toan Ngyen, Minh~Nhat Vu, An Vuong, Dzung Nguyen, Thieu Vo, Ngan Le, and Anh Nguyen.
\newblock Open-vocabulary affordance detection in 3d point clouds.
\newblock \emph{arXiv}, 2023.

\bibitem[Nichol and Dhariwal(2021)]{nichol2021improved}
Alexander~Quinn Nichol and Prafulla Dhariwal.
\newblock Improved denoising diffusion probabilistic models.
\newblock In \emph{ICML}, 2021.

\bibitem[OpenAI(2023)]{chatgpt}
OpenAI.
\newblock Chatgpt.
\newblock \emph{https://chat.openai.com}, 2023.

\bibitem[Paszke et~al.(2019)Paszke, Gross, Massa, Lerer, Bradbury, Chanan, Killeen, Lin, Gimelshein, Antiga, et~al.]{paszke2019pytorch}
Adam Paszke, Sam Gross, Francisco Massa, Adam Lerer, James Bradbury, Gregory Chanan, Trevor Killeen, Zeming Lin, Natalia Gimelshein, Luca Antiga, et~al.
\newblock Pytorch: An imperative style, high-performance deep learning library.
\newblock \emph{NeurIPS}, 2019.

\bibitem[Pi et~al.(2023)Pi, Peng, Yang, Zhou, and Bao]{Pi_2023_ICCV}
Huaijin Pi, Sida Peng, Minghui Yang, Xiaowei Zhou, and Hujun Bao.
\newblock Hierarchical generation of human-object interactions with diffusion probabilistic models.
\newblock In \emph{ICCV}, 2023.

\bibitem[Plappert et~al.(2016)Plappert, Mandery, and Asfour]{kit2016}
Matthias Plappert, Christian Mandery, and Tamim Asfour.
\newblock The kit motion-language dataset.
\newblock \emph{Big Data}, 2016.

\bibitem[Qi et~al.(2017)Qi, Yi, Su, and Guibas]{qi2017pointnet++}
Charles~Ruizhongtai Qi, Li Yi, Hao Su, and Leonidas~J Guibas.
\newblock Pointnet++: Deep hierarchical feature learning on point sets in a metric space.
\newblock \emph{NeurIPS}, 2017.

\bibitem[Ramesh et~al.(2021)Ramesh, Pavlov, Goh, Gray, Voss, Radford, Chen, and Sutskever]{ramesh2021zero}
Aditya Ramesh, Mikhail Pavlov, Gabriel Goh, Scott Gray, Chelsea Voss, Alec Radford, Mark Chen, and Ilya Sutskever.
\newblock Zero-shot text-to-image generation.
\newblock In \emph{ICML}, 2021.

\bibitem[Razali and Demiris(2023)]{razali2023action}
Haziq Razali and Yiannis Demiris.
\newblock Action-conditioned generation of bimanual object manipulation sequences.
\newblock In \emph{AAAI}, 2023.

\bibitem[Reid et~al.(2024)Reid, Savinov, Teplyashin, Lepikhin, Lillicrap, Alayrac, Soricut, Lazaridou, Firat, Schrittwieser, et~al.]{reid2024gemini}
Machel Reid, Nikolay Savinov, Denis Teplyashin, Dmitry Lepikhin, Timothy Lillicrap, Jean-baptiste Alayrac, Radu Soricut, Angeliki Lazaridou, Orhan Firat, Julian Schrittwieser, et~al.
\newblock Gemini 1.5: Unlocking multimodal understanding across millions of tokens of context.
\newblock \emph{arXiv preprint arXiv:2403.05530}, 2024.

\bibitem[Rempe et~al.(2023)Rempe, Luo, Peng, Yuan, Kitani, Kreis, Fidler, and Litany]{rempeluo2023tracepace}
Davis Rempe, Zhengyi Luo, Xue~Bin Peng, Ye Yuan, Kris Kitani, Karsten Kreis, Sanja Fidler, and Or Litany.
\newblock Trace and pace: Controllable pedestrian animation via guided trajectory diffusion.
\newblock In \emph{CVPR}, 2023.

\bibitem[Rombach et~al.(2022)Rombach, Blattmann, Lorenz, Esser, and Ommer]{rombach2022high}
Robin Rombach, Andreas Blattmann, Dominik Lorenz, Patrick Esser, and Bj{\"o}rn Ommer.
\newblock High-resolution image synthesis with latent diffusion models.
\newblock In \emph{CVPR}, 2022.

\bibitem[Saharia et~al.(2022)Saharia, Chan, Saxena, Li, Whang, Denton, Ghasemipour, Gontijo~Lopes, Karagol~Ayan, Salimans, et~al.]{saharia2022photorealistic}
Chitwan Saharia, William Chan, Saurabh Saxena, Lala Li, Jay Whang, Emily~L Denton, Kamyar Ghasemipour, Raphael Gontijo~Lopes, Burcu Karagol~Ayan, Tim Salimans, et~al.
\newblock Photorealistic text-to-image diffusion models with deep language understanding.
\newblock In \emph{NeurIPS}, 2022.

\bibitem[Shafir et~al.(2023)Shafir, Tevet, Kapon, and Bermano]{shafir2023human}
Yonatan Shafir, Guy Tevet, Roy Kapon, and Amit~H. Bermano.
\newblock Human motion diffusion as a generative prior.
\newblock \emph{arXiv}, 2023.

\bibitem[Song et~al.(2024)Song, Zhang, Li, Gao, Hao, Hou, Chen, Li, and Qin]{song2024hoianimator}
Wenfeng Song, Xinyu Zhang, Shuai Li, Yang Gao, Aimin Hao, Xia Hou, Chenglizhao Chen, Ning Li, and Hong Qin.
\newblock Hoianimator: Generating text-prompt human-object animations using novel perceptive diffusion models.
\newblock In \emph{Proceedings of the IEEE/CVF Conference on Computer Vision and Pattern Recognition}, pages 811--820, 2024.

\bibitem[Starke et~al.(2019)Starke, Zhang, Komura, and Saito]{starke2019neural}
Sebastian Starke, He Zhang, Taku Komura, and Jun Saito.
\newblock Neural state machine for character-scene interactions.
\newblock \emph{TOG}, 2019.

\bibitem[Sun and Chowdhary(2023)]{sun2023towards}
Jiarui Sun and Girish Chowdhary.
\newblock Towards globally consistent stochastic human motion prediction via motion diffusion.
\newblock \emph{arXiv}, 2023.

\bibitem[Taheri et~al.(2022)Taheri, Choutas, Black, and Tzionas]{taheri2021goal}
Omid Taheri, Vasileios Choutas, Michael~J. Black, and Dimitrios Tzionas.
\newblock {GOAL}: {G}enerating {4D} whole-body motion for hand-object grasping.
\newblock In \emph{CVPR}, 2022.

\bibitem[Team et~al.(2024)Team, Riviere, Pathak, Sessa, Hardin, Bhupatiraju, Hussenot, Mesnard, Shahriari, Ram{\'e}, et~al.]{team2024gemma}
Gemma Team, Morgane Riviere, Shreya Pathak, Pier~Giuseppe Sessa, Cassidy Hardin, Surya Bhupatiraju, L{\'e}onard Hussenot, Thomas Mesnard, Bobak Shahriari, Alexandre Ram{\'e}, et~al.
\newblock Gemma 2: Improving open language models at a practical size.
\newblock \emph{arXiv preprint arXiv:2408.00118}, 2024.

\bibitem[Tevet et~al.(2023)Tevet, Raab, Gordon, Shafir, Cohen-or, and Bermano]{tevet2023human}
Guy Tevet, Sigal Raab, Brian Gordon, Yonatan Shafir, Daniel Cohen-or, and Amit~Haim Bermano.
\newblock Human motion diffusion model.
\newblock In \emph{ICLR}, 2023.

\bibitem[Tian et~al.(2023)Tian, Zheng, and Liang]{tian2023transfusion}
Sibo Tian, Minghui Zheng, and Xiao Liang.
\newblock Transfusion: A practical and effective transformer-based diffusion model for 3d human motion prediction.
\newblock \emph{arXiv}, 2023.

\bibitem[Touvron et~al.(2023)Touvron, Martin, Stone, Albert, Almahairi, Babaei, Bashlykov, Batra, Bhargava, Bhosale, et~al.]{touvron2023llama}
Hugo Touvron, Louis Martin, Kevin Stone, Peter Albert, Amjad Almahairi, Yasmine Babaei, Nikolay Bashlykov, Soumya Batra, Prajjwal Bhargava, Shruti Bhosale, et~al.
\newblock Llama 2: Open foundation and fine-tuned chat models.
\newblock \emph{arXiv preprint arXiv:2307.09288}, 2023.

\bibitem[Vaswani et~al.(2017)Vaswani, Shazeer, Parmar, Uszkoreit, Jones, Gomez, Kaiser, and Polosukhin]{vaswani2017attention}
Ashish Vaswani, Noam Shazeer, Niki Parmar, Jakob Uszkoreit, Llion Jones, Aidan~N Gomez, {\L}ukasz Kaiser, and Illia Polosukhin.
\newblock Attention is all you need.
\newblock \emph{NeurIPS}, 2017.

\bibitem[Wang et~al.(2022{\natexlab{a}})Wang, Rong, Liu, Yan, Lin, and Dai]{wang2022towards}
Jingbo Wang, Yu Rong, Jingyuan Liu, Sijie Yan, Dahua Lin, and Bo Dai.
\newblock Towards diverse and natural scene-aware 3d human motion synthesis.
\newblock In \emph{CVPR}, 2022{\natexlab{a}}.

\bibitem[Wang et~al.(2023)Wang, Lin, Zeng, Luo, Zhang, and Zhang]{wang2023physhoi}
Yinhuai Wang, Jing Lin, Ailing Zeng, Zhengyi Luo, Jian Zhang, and Lei Zhang.
\newblock Physhoi: Physics-based imitation of dynamic human-object interaction.
\newblock \emph{arXiv preprint arXiv:2312.04393}, 2023.

\bibitem[Wang et~al.(2022{\natexlab{b}})Wang, Chen, Liu, Zhu, Liang, and Huang]{wang2022humanise}
Zan Wang, Yixin Chen, Tengyu Liu, Yixin Zhu, Wei Liang, and Siyuan Huang.
\newblock Humanise: Language-conditioned human motion generation in 3d scenes.
\newblock \emph{NeurIPS}, 2022{\natexlab{b}}.

\bibitem[Wei et~al.(2023)Wei, Sun, Sun, Li, Hu, Li, and Lu]{wei2023understanding}
Dong Wei, Xiaoning Sun, Huaijiang Sun, Bin Li, Shengxiang Hu, Weiqing Li, and Jianfeng Lu.
\newblock Understanding text-driven motion synthesis with keyframe collaboration via diffusion models.
\newblock \emph{arXiv}, 2023.

\bibitem[Wu et~al.(2022)Wu, Wang, Zhang, Zhang, Hilliges, Yu, and Tang]{wu2022saga}
Yan Wu, Jiahao Wang, Yan Zhang, Siwei Zhang, Otmar Hilliges, Fisher Yu, and Siyu Tang.
\newblock Saga: Stochastic whole-body grasping with contact.
\newblock In \emph{ECCV}, 2022.

\bibitem[Xie et~al.(2024)Xie, Jampani, Zhong, Sun, and Jiang]{xie2023omnicontrol}
Yiming Xie, Varun Jampani, Lei Zhong, Deqing Sun, and Huaizu Jiang.
\newblock Omnicontrol: Control any joint at any time for human motion generation.
\newblock In \emph{ICLR}, 2024.

\bibitem[Xu et~al.(2023)Xu, Li, Wang, and Gui]{xu2023interdiff}
Sirui Xu, Zhengyuan Li, Yu-Xiong Wang, and Liang-Yan Gui.
\newblock Interdiff: Generating 3d human-object interactions with physics-informed diffusion.
\newblock In \emph{ICCV}, 2023.

\bibitem[Xu et~al.(2024)Xu, Wang, Wang, and Gui]{xu2024interdreamer}
Sirui Xu, Ziyin Wang, Yu-Xiong Wang, and Liang-Yan Gui.
\newblock Interdreamer: Zero-shot text to 3d dynamic human-object interaction.
\newblock \emph{arXiv preprint arXiv:2403.19652}, 2024.

\bibitem[Yang et~al.(2024)Yang, Niu, Jiang, Zhang, and Huang]{yang2024f}
Jie Yang, Xuesong Niu, Nan Jiang, Ruimao Zhang, and Siyuan Huang.
\newblock F-hoi: Toward fine-grained semantic-aligned 3d human-object interactions.
\newblock \emph{arXiv preprint arXiv:2407.12435}, 2024.

\bibitem[Zhang et~al.(2023{\natexlab{a}})Zhang, Zhang, Cun, Huang, Zhang, Zhao, Lu, and Shen]{zhang2023generating}
Jianrong Zhang, Yangsong Zhang, Xiaodong Cun, Shaoli Huang, Yong Zhang, Hongwei Zhao, Hongtao Lu, and Xi Shen.
\newblock T2m-gpt: Generating human motion from textual descriptions with discrete representations.
\newblock In \emph{CVPR}, 2023{\natexlab{a}}.

\bibitem[Zhang et~al.(2022{\natexlab{a}})Zhang, Cai, Pan, Hong, Guo, Yang, and Liu]{zhang2022motiondiffuse}
Mingyuan Zhang, Zhongang Cai, Liang Pan, Fangzhou Hong, Xinying Guo, Lei Yang, and Ziwei Liu.
\newblock Motiondiffuse: Text-driven human motion generation with diffusion model.
\newblock \emph{arXiv}, 2022{\natexlab{a}}.

\bibitem[Zhang et~al.(2023{\natexlab{b}})Zhang, Guo, Pan, Cai, Hong, Li, Yang, and Liu]{zhang2023remodiffuse}
Mingyuan Zhang, Xinying Guo, Liang Pan, Zhongang Cai, Fangzhou Hong, Huirong Li, Lei Yang, and Ziwei Liu.
\newblock Remodiffuse: Retrieval-augmented motion diffusion model.
\newblock \emph{arXiv}, 2023{\natexlab{b}}.

\bibitem[Zhang et~al.(2022{\natexlab{b}})Zhang, Bhatnagar, Starke, Guzov, and Pons-Moll]{zhang2022couch}
Xiaohan Zhang, Bharat~Lal Bhatnagar, Sebastian Starke, Vladimir Guzov, and Gerard Pons-Moll.
\newblock Couch: Towards controllable human-chair interactions.
\newblock In \emph{ECCV}, 2022{\natexlab{b}}.

\bibitem[Zhang et~al.(2023{\natexlab{c}})Zhang, Liu, Aberman, and Hanocka]{zhang2023tedi}
Zihan Zhang, Richard Liu, Kfir Aberman, and Rana Hanocka.
\newblock Tedi: Temporally-entangled diffusion for long-term motion synthesis.
\newblock \emph{arXiv}, 2023{\natexlab{c}}.

\bibitem[Zhao et~al.(2023)Zhao, Zhang, Wang, Beeler, and Tang]{zhao2023synthesizing}
Kaifeng Zhao, Yan Zhang, Shaofei Wang, Thabo Beeler, and Siyu Tang.
\newblock Synthesizing diverse human motions in 3d indoor scenes.
\newblock \emph{arXiv}, 2023.

\end{thebibliography}
}

\clearpage
\appendix
\setcounter{page}{1}
\maketitlesupplementary


\section{Additional Details of Methodology}
\label{supp:method}
In~\sref{sec:method} of our main paper, we presented the foundational design of each key component in our HOI-Diff pipeline. Here, we delve into an elaborate explanation of model architecture, learning objectives and additional details associated with each crucial component.

\subsection{Dual-branch diffusion model (DBDM)} 
\label{supp:dbdm}
The Communication Module (CM) in DBDM is based on the cross attention mechanism. Formally,
\begin{equation}
    \tilde{\boldsymbol f}^h = \operatorname{MLP}(\operatorname{Attn}(\vf^h \rmW_Q, \vf^o \rmW_K, \vf^o \rmW_V)),
\label{eq:com_module1}
\end{equation}
\begin{equation}
     \tilde{\boldsymbol f}^o = \operatorname{MLP}(\operatorname{Attn}(\vf^o \rmW_Q, \vf^h \rmW_K, \vf^h \rmW_V)), 
\label{eq:com_module2}
\end{equation}
where $\operatorname{MLP}(\cdot)$ denotes fully-connected layers, $\operatorname{Attn}(\cdot)$ is the attention block~\citep{vaswani2017attention}, and $\rmW_Q, \rmW_K, \rmW_V$ are learned projection matrices for query, key, and value, respectively. 

The training objective of this full model is based on reconstruction loss
\begin{align}\label{eq:rec_hoi}
\begin{split}
    \mathcal{L}_{hoi} = \E_{t \sim [1, T]} \|M_{\theta}(\boldsymbol x_t, t,  \boldsymbol c) - \boldsymbol x_0\|^2_2,
\end{split}
\end{align}
where $\boldsymbol x_0$ is the ground truth of the HOI sequence.

\subsection{Affordance prediction diffusion model (APDM).}
\label{supp:apdm}
\myparagraph{Model architecture.}
The affordance prediction diffusion model
comprises eight Transformer layers for the encoder with a PointNet++~\citep{qi2017pointnet++} to encode the object's point clouds. The training objective of this diffusion model is also based on reconstruction loss
\begin{align}\label{eq:rec_aff}
\begin{split}
    \mathcal{L}_{aff} = \E_{t \sim [1, T]} \|{A}_{\theta}(\boldsymbol y_t, t, \boldsymbol p, \boldsymbol d) - \boldsymbol y_0\|^2_2,
\end{split}
\end{align}
where $\boldsymbol y_0$ is the ground-truth affordance data. $ \boldsymbol p$ and $\boldsymbol d$ denote object point cloud and text description (prompt), respectively. $A_{\theta}$ represents the affordance prediction diffusion model.

\begin{figure}[h]
    \centering
    \includegraphics[width=0.75\linewidth]{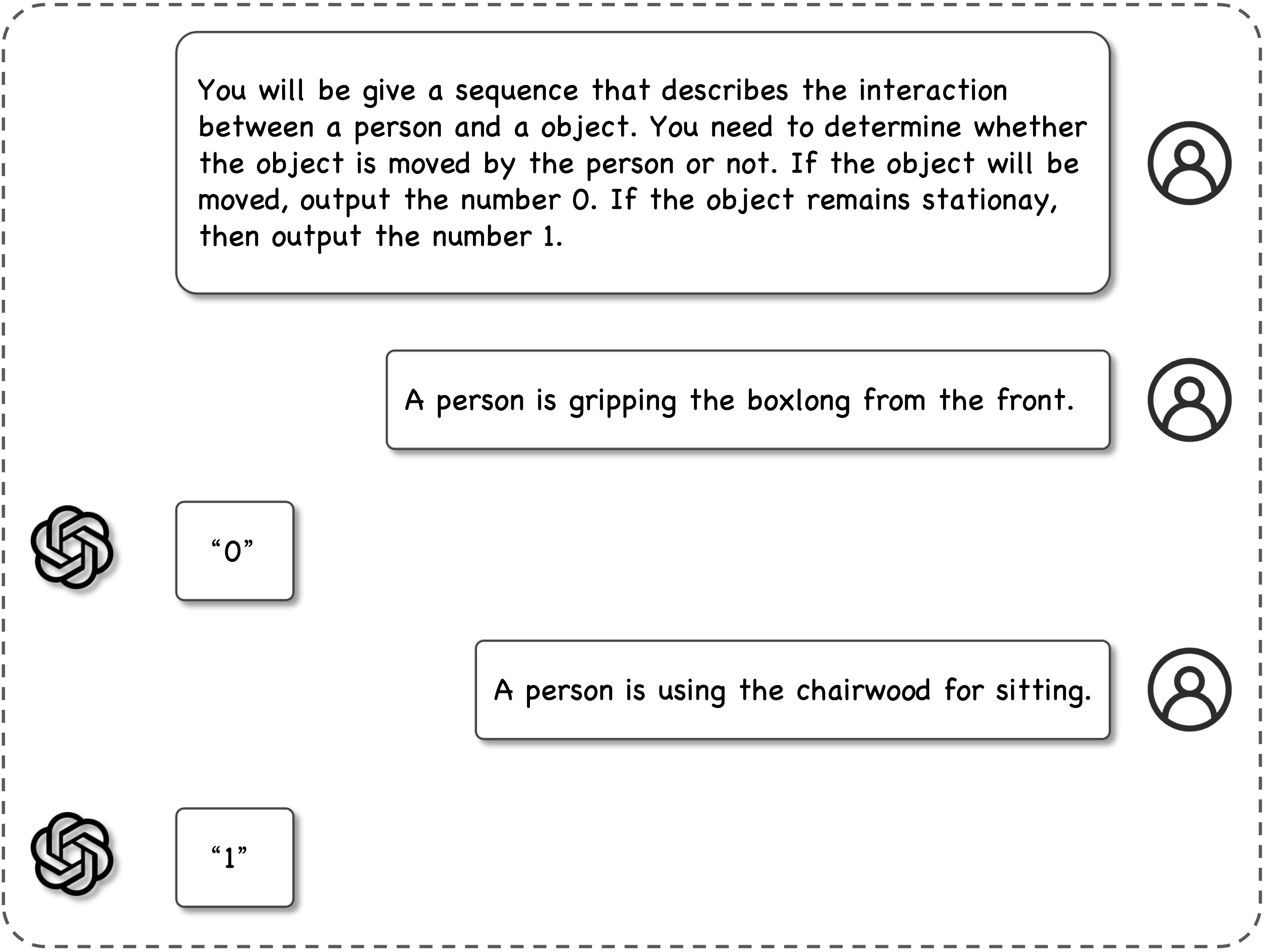}
    \caption{
    \textbf{Prompt template for inferring object state.}
    }
    \label{fig:prompt_template}
\end{figure}

\myparagraph{Inferring object state with GPT-3.5-turbo in APDM.} To infer the state of an object, we directly leverage the strong prior knowledge of large language models to derive the result. Specifically, we utilize the GPT-3.5-turbo~\citep{chatgpt} API by inputting specific instructions, allowing it to infer the result directly based on the input HOI text description. The prompt template for instruction is shown in~\Fref{fig:prompt_template}.

\subsection{Affordance-guided interaction correction.}
\label{supp:interaction_correction}
During the inference stage, it's found that the predicted object contact positions may occasionally be inaccurately positioned, residing either inside or outside the object. To rectify this, we implement post-processing steps that replace these predicted contact points, denoted as $\boldsymbol y_0^{o}$, with their nearest neighbors from the object's point clouds. This adjustment aims to enhance the accuracy of the updated contact points, aligning them more closely with their actual positions on the object's surface. However, employing these updated contact points directly for contact constraints, particularly in the absence of detailed human shape information, introduces a new challenge. It can potentially lead to penetration issues within the contact area while reconstructing the human mesh in the final stage. To mitigate contact penetration, we adopt a method that recalculates points at a specified distance outward, perpendicular to the normal, originating from the object's contact points. This process can formulated as: $\tilde{\boldsymbol y}{0}^{o} = \hat{\boldsymbol y}_{0}^{o} + v_n^{i} * d$, where $i \in \{1, 2\}$ indicates the $i^{th}$ object contact points, $v_n^i$ denotes the normal vector at that point and $d=0.05$ is a contact distance threshold. 

As for smoothness term, we formulate it as 
\begin{align}\label{eq:guide_smo}
\begin{split}
\small{
G_{smo} =  \sum_{l=1}^{L-1}  \norm{\vx_0^o(l+1)  - \vx_0^o(l) }^2,
}
\end{split}
\end{align}
where $\vx_0^o(l)$ is the predicted 6DoF pose of the object in the $l$-th frame.

\newcommand{\comment}[1]{\textcolor{gray}{#1}}

\begin{center}
\scalebox{0.75}{
\begin{minipage}{1.2\linewidth}
\begin{algorithm}[H]
\caption{Affordance-guided Interaction Correction }\label{alg:guidance}
\renewcommand\arraystretch{0.8}
\begin{algorithmic}[1]
\Require Input $\vc=(\vd, \vp)$ consisting of a textual description $\vd$ and object point cloud $\vp$, \shortname~model $\textbf{M}_{\theta}$, objective function $G(\vmu^h_t, \vmu_t^o, \boldsymbol y_0)$, and estimated affordance $\vy_0 = (\vy_0^{h}, \vy_0^{o}, \vy_0^{s})$. 
\State $\vx^{h}_T, \vx^{o}_T \leftarrow \text{sample from } \mathcal{N}(\mathbf{0}, \mathbf{I})$
\State $K$ = 1
\ForAll{$t$ from $T$ to $1$}
    \State $\vx_0^{h}, \vx_0^{o} \leftarrow M_{\theta}(\vx_t^h, \vx_t^o, t, \vc)$  \comment{\#  Get $\vmu_t^h, \vmu_t^o$ according to Eq.(\ref{eq:mu}) with  $\Sigma_t$ }
    \If{$t=1$}
        \State $K=100$
    \EndIf
    \ForAll{$k$ from $K$ to $1$}  \comment{\small \# Separately perturb}  
        \State $\vmu^h_t \leftarrow \vmu^h_t  - \tau_1\Sigma_t \nabla_{\vmu_t^h} G(\vmu^h_t, \vmu^o_t, \boldsymbol y_0)$,\quad $\vmu^o_t \leftarrow \vmu^o_t  - \tau_2\Sigma_t \nabla_{\mu_t^o} G(\vmu^h_t, \vmu_t^o, \boldsymbol y_0)$ 
    \EndFor    
    \State $\x_{t-1}^{h} \sim \mathcal{N}(\vmu^h_t, \Sigma_t)$,\quad $\x_{t-1}^{o} \sim \mathcal{N}(\vmu^o_t, \Sigma_t)$ 
\EndFor\\
\Return $\vx_{0}^{h}, \vx_{0}^{o}$
\end{algorithmic}
\end{algorithm}
\end{minipage}
}
\end{center}

\section{Implementation Details}
\label{supp:implementation}
Both our DBDM and APDM are built on the Transformer \citep{vaswani2017attention} architecture. Similar to MDM \citep{tevet2023human}, we employ the CLIP model to encode text prompts, adhering to a classifier-free generation process. Our models are trained using PyTorch \citep{paszke2019pytorch} on 1 NVIDIA A5000 GPU. We set control strength of guidance as $\tau_1$ = 1, $\tau_2$ = 100, and  ${\Sigma}_t = min(\Sigma_t, 0.01)$. Both the DBDM and APDM are trained on the same data for 20k steps. 

Both the DBDM and APDM architectures of HOI-Diff are based on Transformers with 4 attention heads, a latent dimension of 512, a dropout of 0.1, a feed-forward size of 1024, and the GeLU activation~\citep{hendrycks2016gaussian}. The number of learned parameters for each model is stated in \Tref{table:learning_param}.


Our training setting involves 20k iterations for the DBDM and 10k iterations for the APDM model. These iterations utilize a batch size of 32 and employ the AdamW optimizer~\citep{loshchilov2017decoupled} with a learning rate set at 10$^{-4}$. We use $T$=1000 and $N$=500 diffusion steps in DBDM and APDM, respectively.

\section{Additional Details of Baselines}
\label{supp:baselines}
\begin{itemize}
\item  MDM$^{finetuned}$: We finetune MDM~\citep{tevet2023human} on BEHAVE dataset without considering the object motion.
\item  MDM*: We extend the original feature dimensions of the input and output processing in MDM~\citep{tevet2023human} from $D^h$ to $D^{h} + D^{o}$, enabling support for HOIs sequences. 
The model is trained from scratch on BEHAVE dataset~\citep{bhatnagar22behave}.

\item PriorMDM*: The proposed approach for dual-person motion generation employs paired fixed MDMs~\citep{tevet2023human} per individual to ensure uniformity within generated human motion distributions. This design leverages a singular ComMDM to coordinate between the two branches of fixed MDM instances, streamlining training and maintaining consistency across generated motions. Given that both branches are based on MDM that pretrained on human motion datasets, direct utilization of them for human-object interactions in our task is infeasible. We maintain one branch dedicated to humans, leveraging pre-trained weights, while adapting the input and output processing of another branch specifically for generating object motion. Following this, we fine-tune the human MDM branch while initiating the learning of object motion from scratch within the object branch. Eventually, we integrate ComMDM to facilitate communication and coordination between these distinct branches handling human and object interactions.

\item InterDiff: InterDiff~\citep{xu2023interdiff} is originally designed for a prediction task rather than text-driven HOIs generation. To tailor it to our task, we replace its Transformer encoder with a CLIP encoder and modify its feature dimensions of the input and output layers.

\end{itemize}

To ensure fair comparisons, all the above baselines as well as our own models are all trained on BEHAVE and OMOMO datasets for 20k steps.

\begin{figure}[t]
    \centering
    \includegraphics[width=1.0\linewidth]{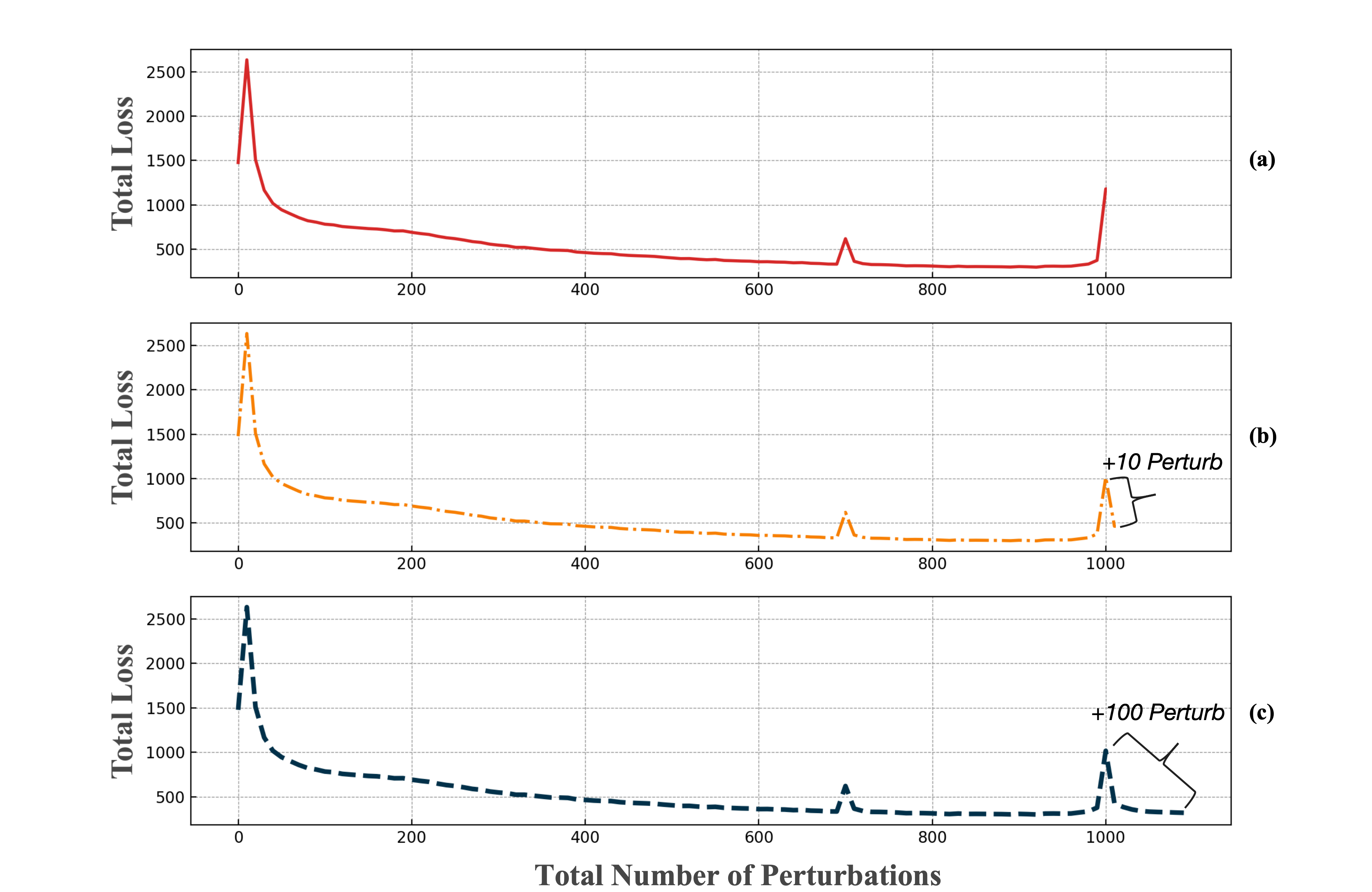}
    \caption{\textbf{Effect of different total numbers of perturbations in the whole denoising process.} (a) Perturb one time in each denoising step (in total $T = 1000$). (b) Perturb one time in first $T-1$ denoising steps, and repeatedly perturb 10 times in the final denoising step. (c) Perturb one time in first $T-1$ denoising steps, and repeatedly perturb 100 times in the final denoising step.}
    \label{fig:perturb}
\end{figure}

\section{Additional Details of Evaluation Metrics}
\label{supp:metrics}
For detailed information regarding metrics employed in human motion generation, including \textit{FID}, \textit{R-Precision}, and \textit{Diversity}, we refer readers to \cite{tevet2023human, Guo_2022_CVPR} for comprehensive understanding. 

\myparagraph{Contact Distance.} Expanding on the concept of \textit{Contact Distance}, we utilize the \textit{chamfer distance} metric to quantify the closeness between human body joints and the object surface. This computation leverages ground-truth affordance data that includes human contact labels and object contact points,
\begin{equation}\label{eq:metric_contact}
    ContactDistance =  \frac{1}{L}\sum_{l}^{L} {CD(\hat{\boldsymbol x}_l^{h}, \hat{\boldsymbol p}_l)},
\end{equation}
where $\hat{\boldsymbol x}_l^{h}$ represents two human contact joints at the $l$-th frame, indexed according to ground-truth contact labels. Additionally, $ \hat{\boldsymbol p}_l$ denotes two object contact points derived from the object motion $\boldsymbol x^o_l$ at frame $l$, also indexed based on ground-truth information. $CD$ denotes the \textit{chamfer distance}.

 \myparagraph{Penetration Score.} We followed the ~\citet{li2023controllable} to compute the penetration score (Pene), each vertex of the body ($V_i$) is queried against the precomputed Signed Distance Field (SDF) of the object. This process yields a corresponding distance value for each vertex. The penetration score is then formalized as:
\begin{equation}
    Pene = \frac{1}{n}\sum_{i=1}^{n}{|min(d_i,0)|},
\end{equation}
measured in centimeters (cm).

\newfloatcommand{capbtabbox}{table}[][]
\vspace{-1.0em}
\begin{figure*}[h!]
    \begin{floatrow}
    \capbtabbox{
        \resizebox{0.4\textwidth}{!}{
            \begin{tabular}{ccc}
            \toprule
            Model &  DBDM &   APDM   \\\hline
            Parameters (·10$^6$)      & 8.82     & 38.92     \\ 
            \bottomrule
            \end{tabular}
        }
    }
    {
      \caption{\textbf{Model Parameters.} The number of learned parameters of our two core architectures.}
      \label{table:learning_param}
    }
    \hspace{-0.5em}
    \capbtabbox{
      \resizebox{0.5\textwidth}{!}{
        \begin{tabular}{c|ccc}
        \toprule
        Method & MDM* &  PriorMDM* & Ours (Full) \\ \hline
        
        Time (s)& 32.3 & 38.6 & 118.0  \\ \hline
        Component &  APDM & DBDM & Interaction Correction \\ \hline
        
        Time (s) & 24.2 & 46.4 & 47.4 \\
        \bottomrule
        \end{tabular}
        }
    }
    {
      \caption{\textbf{Inference Time (on NVIDIA A5000 GPU).} We report the inference time for baselines, our full method, and its key components. }
      \label{table:inference_time}
    }
    \end{floatrow}
\end{figure*}

\section{Inference Time}
\label{supp:inference}
In \Tref{table:inference_time}, we provide the inference times for both baselines and our full method, including its key components. All measurements were conducted using an NVIDIA A5000 GPU. Training an additional model for affordance information and using classifier guidance for interaction correction do contribute to increased inference costs. However, despite the longer inference time, our complete method notably enhances the accuracy of 3D HOIs generation.




\begin{table}[h]
\centering
\resizebox{0.8\linewidth}{!}{
\begin{tabular}{l|c|c|c}
    \toprule
    & Params (M) & FID $\downarrow$ & R-precision (Top-3) $\uparrow$\\ 
     \midrule
    MDM$^{*}$ & 49.85  & 6.98 & 0.36\\
    Ours (Full) & 47.74 & 1.62 & 0.46 \\
    \bottomrule
\end{tabular}
}
\caption{With comparable model size, the performance results of MDM$^{*}$ and Ours (Full).}
\label{table:model_params}
\end{table}

\begin{figure}[t]
    \centering
    \includegraphics[width=0.75\linewidth]{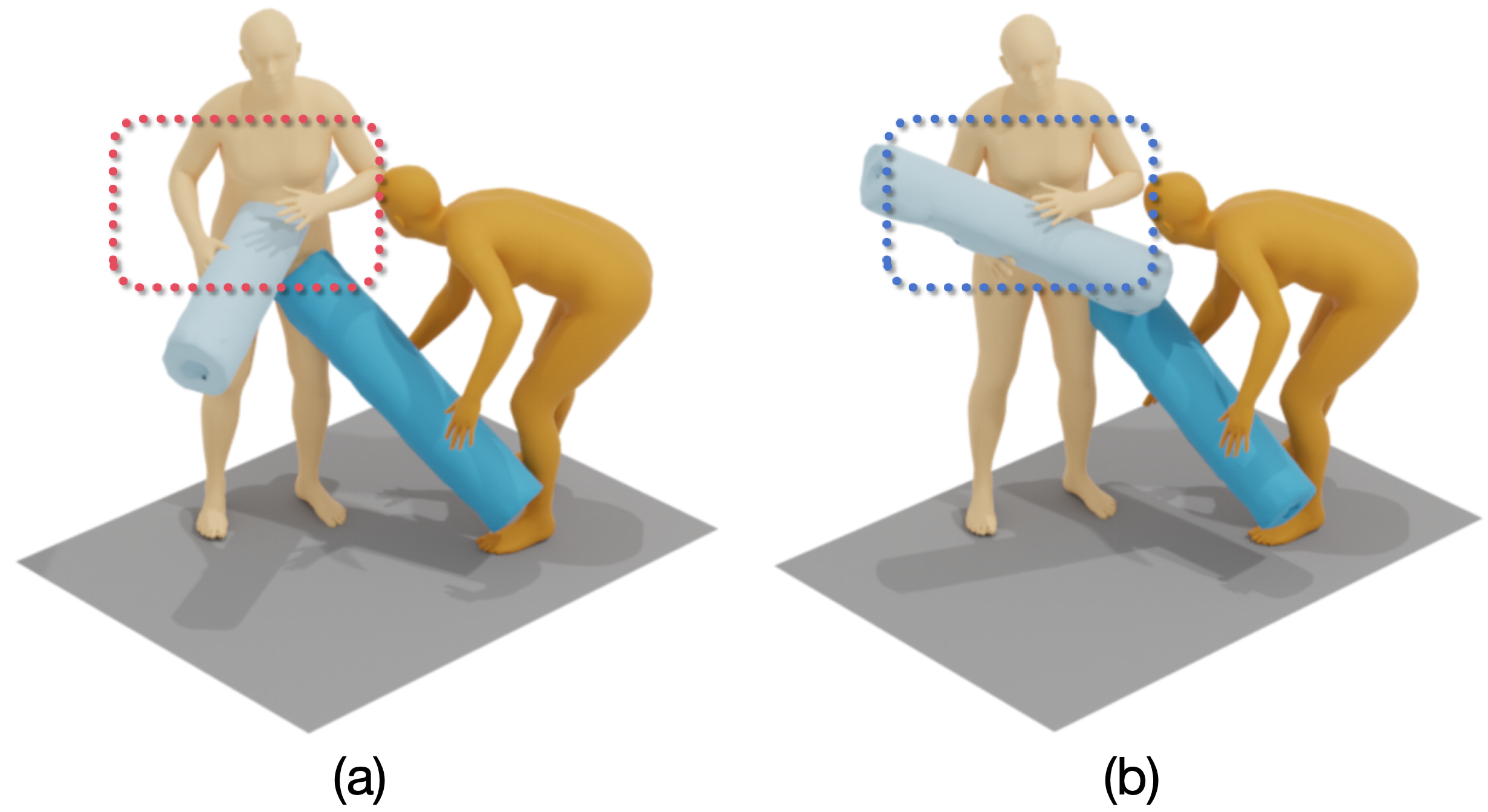}
    \caption{\textbf{Effect of different control strengths for classifier guidance.} (a) We use equal strengths of $\tau_1=1, \tau_2=1$ to perturb the predicted mean of human motion and object motion, respectively. (b) We use different strengths of $\tau_1=1, \tau_2=100$ for the perturbation. We can see that different strengths work better.
    }
    \label{fig:different_strength}
\end{figure}

\section{Additional Ablation Studies}
\label{supp:ablation}


\myparagraph{Different perturbing times in classifier guidance.} As discussed in \sref{method:guidance}, in the later stage of classifier guidance, diffusion models tend to strongly attenuate the introduced signals. Therefore, we iteratively perturb the predicted mean of motion for $K$ times at the final denoising step. In \Fref{fig:perturb}, we present the ablation results, illustrating the impact of different numbers of perturbations. Notably, we observe that employing 100 perturbations leads to re-convergence and yields the desired results.

\myparagraph{Different guidance strength.} As detailed in~\sref{method:guidance}, we employ distinct control strengths for classifier guidance, considering the varying feature densities in predicted human and object motion. Rather than employing equal control strengths, we opt to assign a higher control strength to object motion, allowing it to closely align with human contact joints, as illustrated in \Fref{fig:different_strength}.

\myparagraph{Different model with comparable model size.} Although our method involves a slightly larger number of model parameters, our model is specifically designed for HOI generation. As seen in the ~\Tref{table:model_params}, if we attempt to scale MDM* to the same model size, its performance remains subpar.

\section{User Study} 
\label{supp:user_study}
\begin{figure*}[t]
    \vspace{-8mm}
    \begin{floatrow}
    \ffigbox{
      \includegraphics[width=0.8\linewidth]{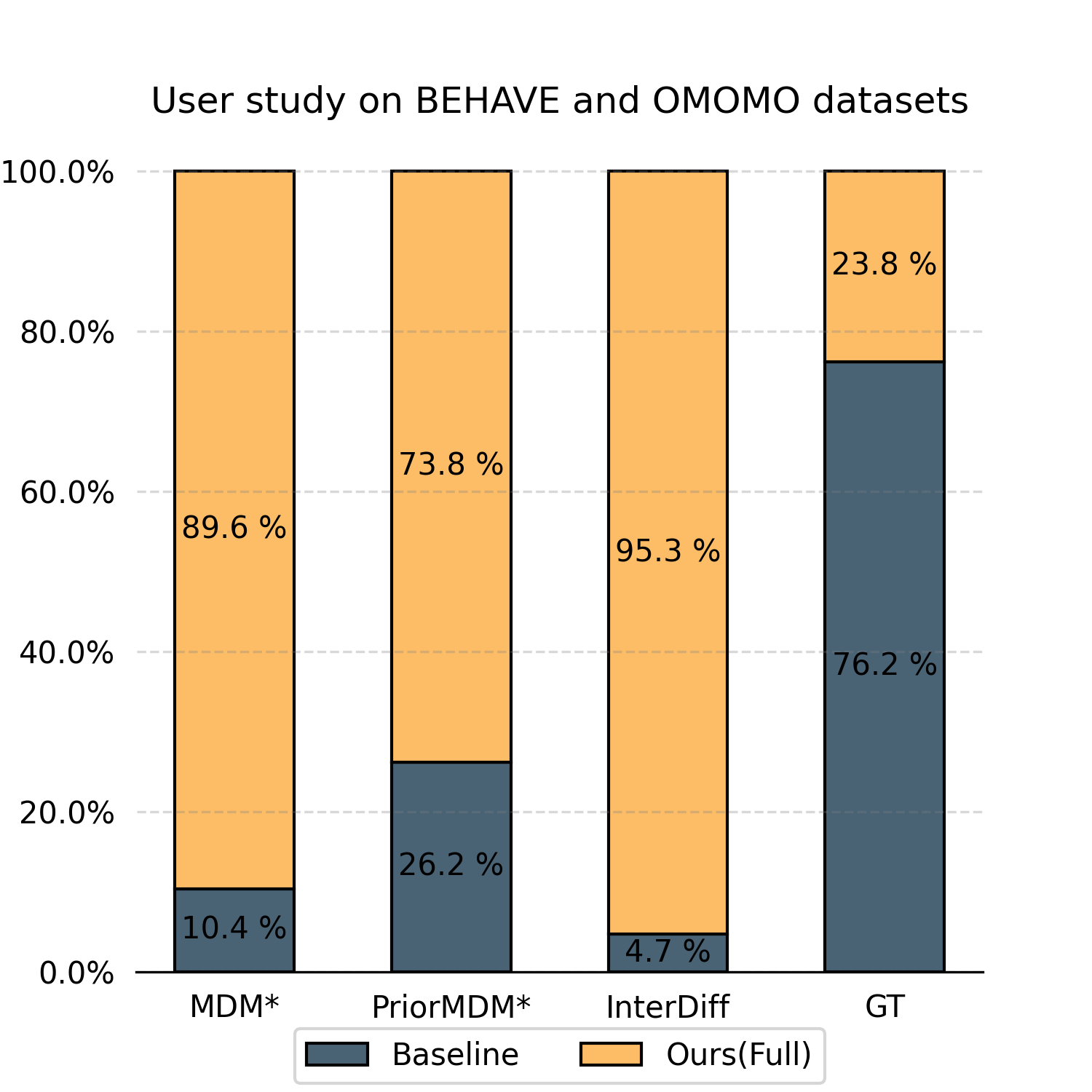}
    }{
      \vspace{-2mm}
      \caption{\textbf{Perceptual User Study.} Most participants prefer our method over the baselines.
      }
      \label{fig:user_study}
    }
    \ffigbox{
        \includegraphics[width=0.8\linewidth]{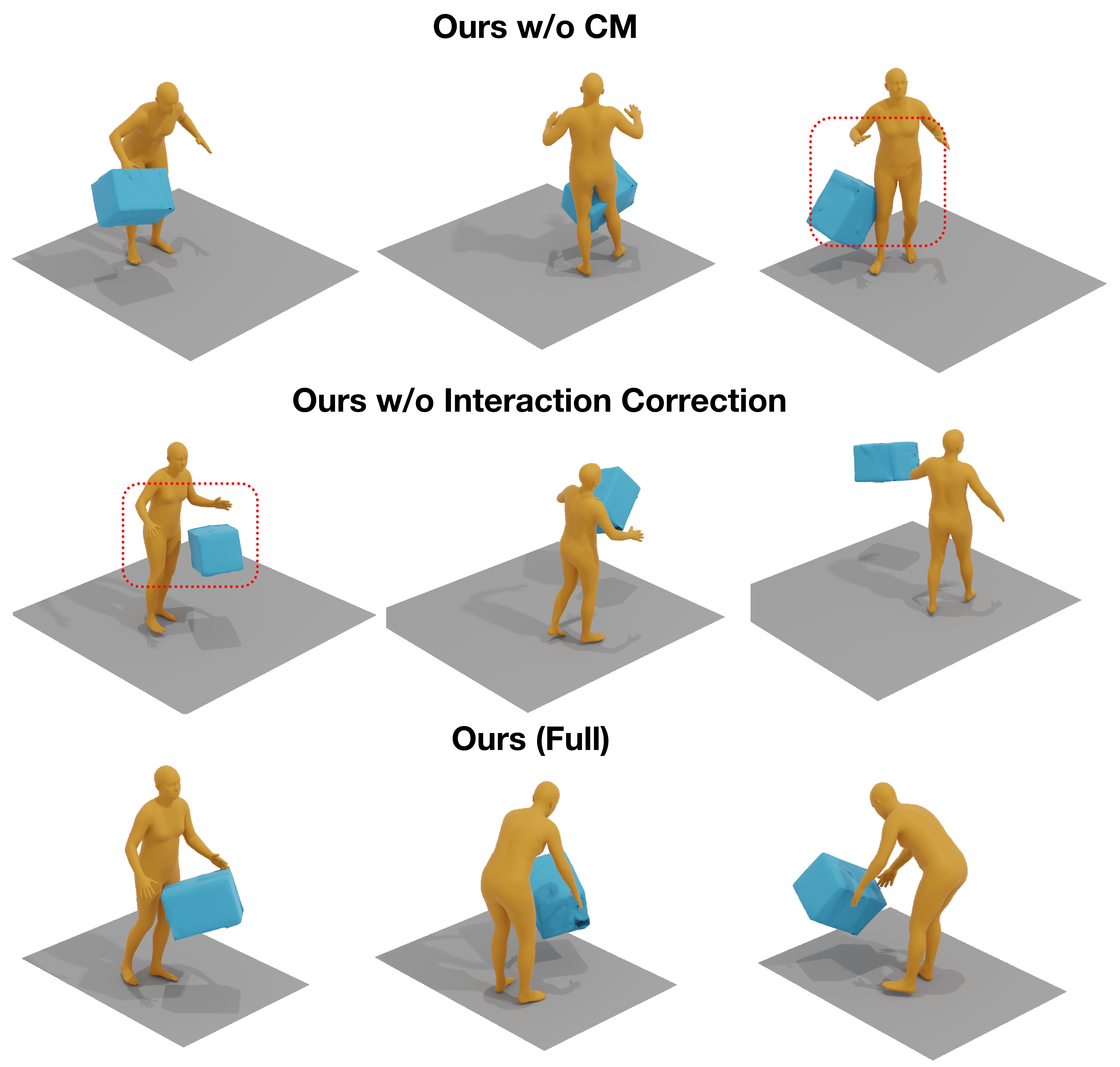}
    }{
      \vspace{-2mm}
      \caption{\textbf{Visual results of different variants of our model in ablation studies.}}
       \label{fig:ablation_vis}
    }
    \end{floatrow}
\end{figure*}

\new{For each method, we select 15 prompts from the BEHAVE dataset and 10 prompts from the OMOMO dataset, covering various interaction types and object items. We sample twice with each prompt to gather a total of 50 results. 40 participants are asked to choose their most preferred generation results from these samples. This user study requires pairwise comparisons of our method with other baseline on generated interaction quality, as shown in \Fref{fig:user_study_quest}. 
The results in \Fref{fig:user_study} indicate strong preference for our method: it is favored over the baselines in 89.6\% (Ours \emph{vs.} MDM*), 73.8\% (Ours \emph{vs.} PriorMDM*) and 95.3\% (Ours \emph{vs.} Interdiff).
}

\section{Additional Qualitative Results}
\label{supp:qualitative}
In this section, we present additional qualitative results showcasing the model's performance evaluated on the OMOMO dataset, and the effectiveness of APDM.

\begin{figure*}[h]
  \centering
    \includegraphics[width=1.0\linewidth]{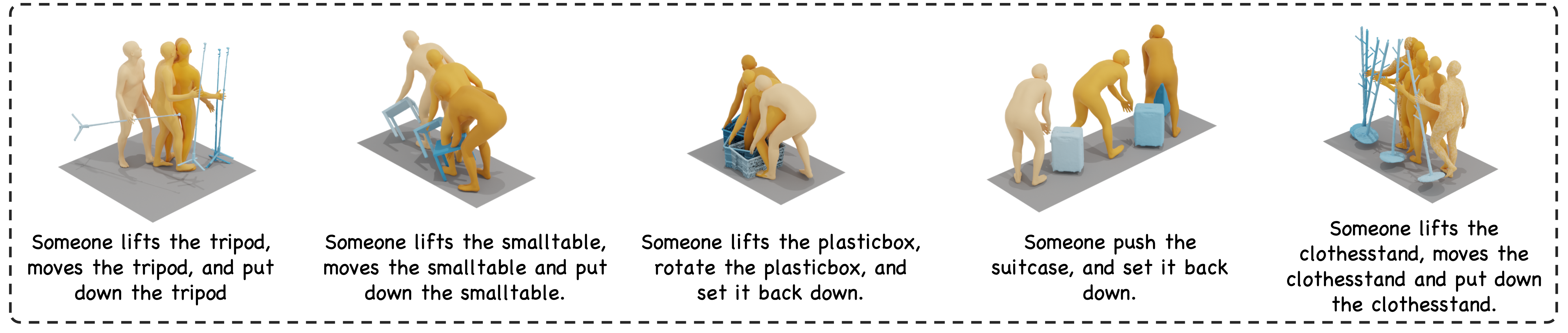}
    \caption{Additional qualitative evaluation on OMOMO dataset. Given
object geometry and text description, our method can generate high-quality human-object interactions even for the unseen objects (tripod, smalltable, suitcase).}
    \label{fig:vis_omomo}
\end{figure*}

\begin{figure}[h]
  \centering
    \includegraphics[width=1.0\linewidth]{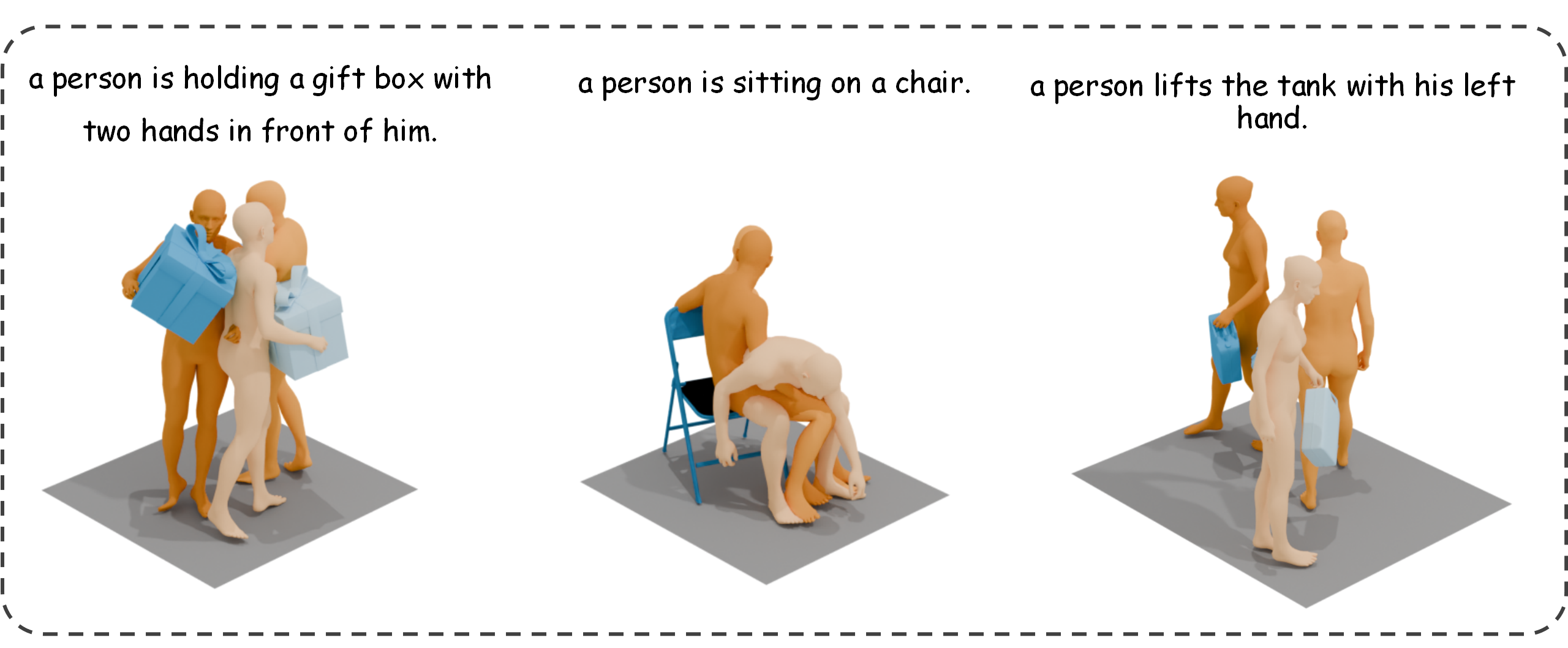}
    \caption{Additional qualitative evaluation on unseen objects.}
    \label{fig:unseen_obj}
\end{figure}

\begin{figure}[t]
  \centering
   \includegraphics[width=0.85\linewidth]{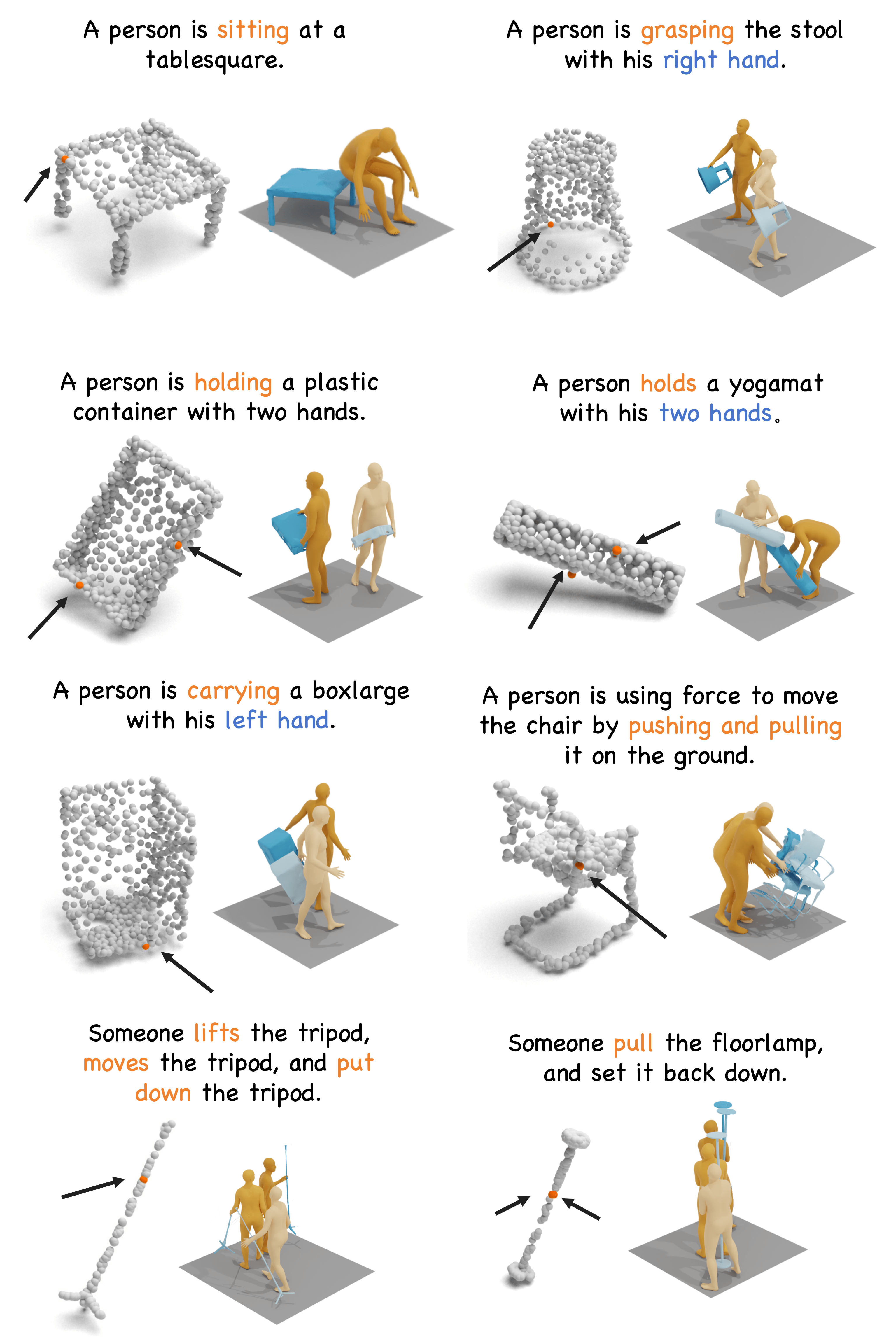}
    \caption{Visual results of estimated contact points. Our APDM, trained on the BEHAVE dataset, can accurately estimating contact positions for objects based on textual descriptions. Furthermore, it showcases the capability to generalize to unseen objects in the OMOMO dataset, as demonstrated in the last row.}
    \label{fig:vis_apdm}
\end{figure}

\myparagraph{Qualitative results on OMOMO dataset.} 
We present additional qualitative results on the OMOMO dataset, rendered with SMPL~\citep{SMPL:2015} shapes, as shown in~\Fref{fig:vis_omomo}. It is evident that our method can generalizes effectively to unseen objects and produce realistic 3D human-object interactions.

\begin{figure}[h]
  \centering
    \includegraphics[width=1.0\linewidth]{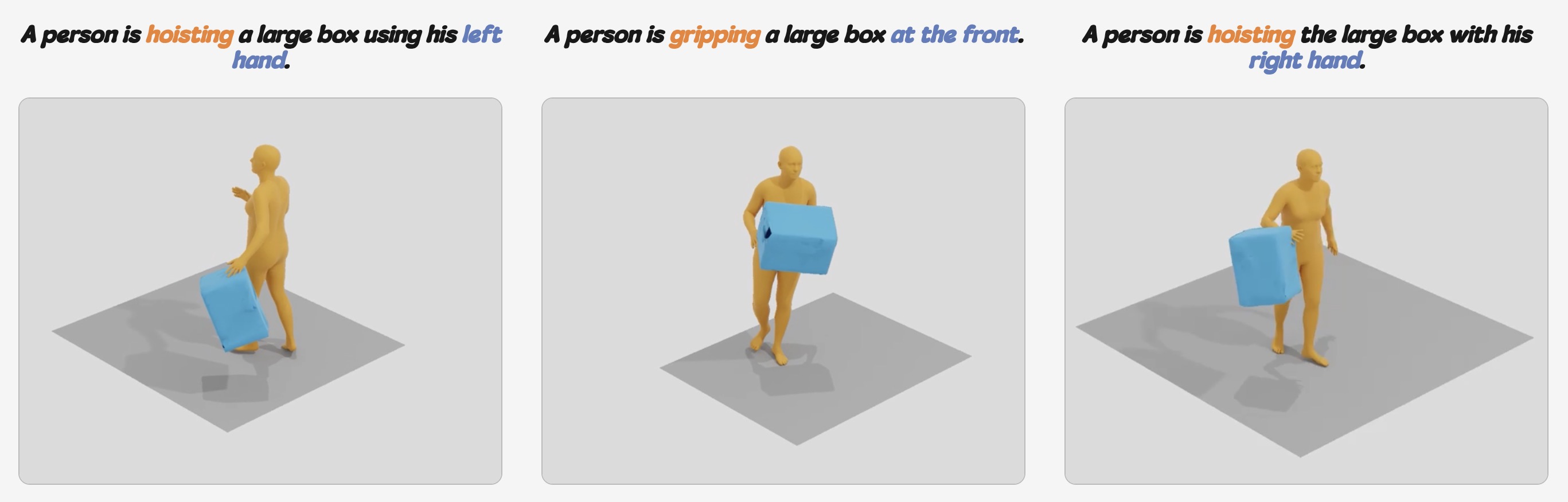}
    \caption{\new{Leveraging the power of the APDM module, our method can generate diverse HOIs for the same object using different contacting body parts and contact points.}}
    \label{fig:different_contact}
\end{figure}

\begin{figure}[h]
    \centering
    \includegraphics[width=0.75\linewidth]{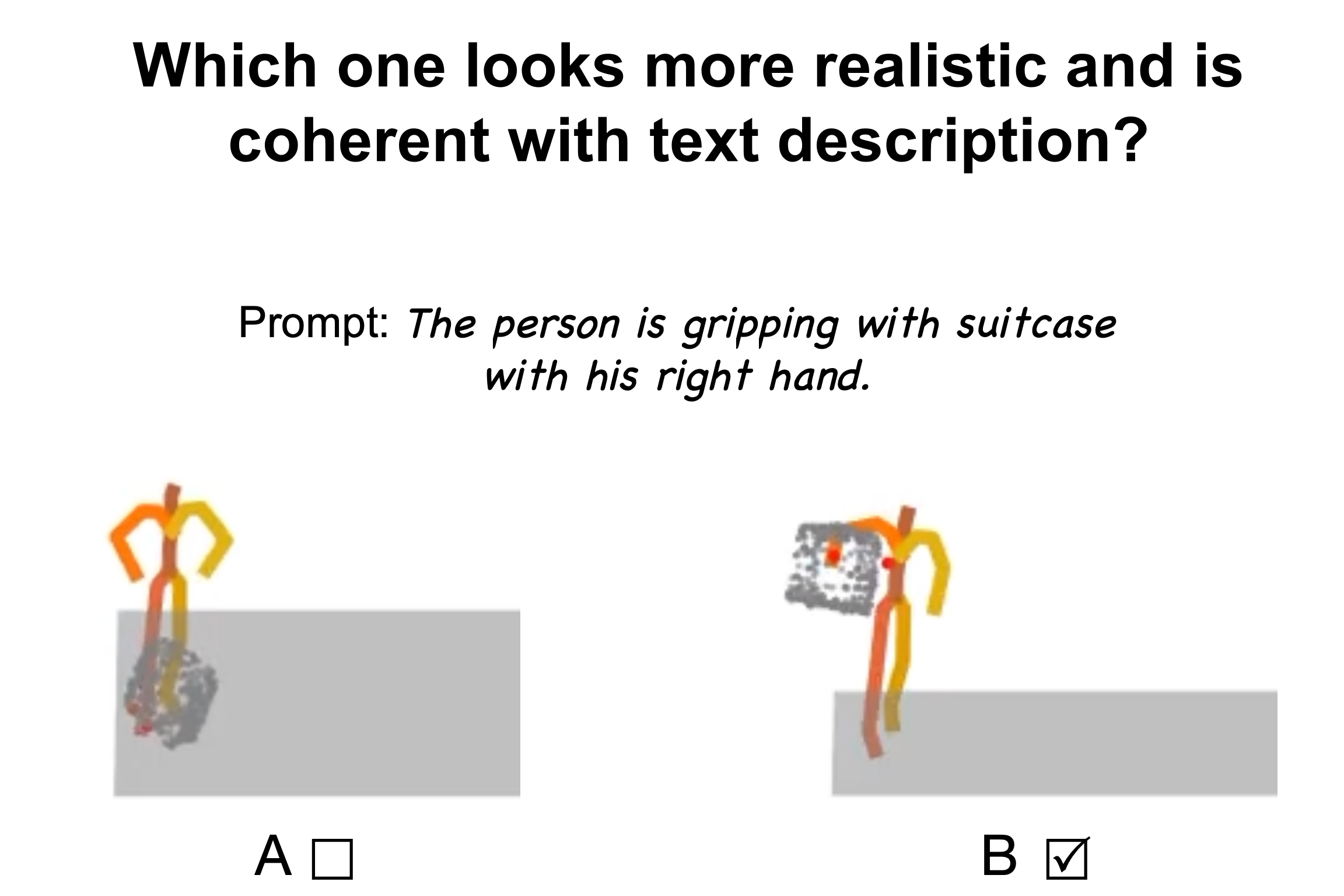}
    \caption{\textbf{An example question for our text-to-hoi user study.}
    }
    \label{fig:user_study_quest}
\end{figure}

\myparagraph{Qualitative results of APDM.} 
To verify the accuracy of estimated contact points on object surface, we provide additional visual results in \Fref{fig:vis_apdm}. It can be seen that our method can predict realistic and practical contact points based on text descriptions. With APDM, we even can generate different interactions with the same object based on the input description, as shown in the~\Fref{fig:different_contact}.

\myparagraph{Generalization capability.} To verify the model's generalization capability, except of unseen object test on OMOMO dataset, we also downloaded several objects from Sketchfab\footnote{https://sketchfab.com/}, adjusted them to a reasonable scale, and used them as inputs. As shown in \Fref{fig:unseen_obj}, our model successfully establishes reasonable HOI contact with these previously unseen objects.

\vspace{-0.8em}
\section{Annotation for BEHAVE Dataset}
\label{supp:dataset}
\myparagraph{Text Annotating Process.}
Initially, we manually annotate the interaction types and the specific human body parts involved, delineating actions like ``lift'' associated with the ``left hand'' or ``hold'' involving ``two hands''. Subsequently, to generate complete sentences, we leverage the capabilities of GPT-3.5 to assist in formulating the entirety of the description. 

\newcolumntype{C}[1]{>{\centering\arraybackslash}m{#1}}
\begin{table}[h]
\begin{center}
\caption{\textbf{Examples of our annotated textual descriptions for the BEHAVE dataset rephrased by GPT-3.5~\citep{chatgpt}.}}
\label{tab:texts}
\vspace{2mm}
\resizebox{0.8\linewidth}{!}{
\begin{tabular}{C{2.2cm} C{5.7cm}}
\hline
\textbf{Object} &  \textbf{Textual Descriptions}  \\
\hline
\multirow{3}{*}{\textit{backpack}} &\small{A person is carrying the backpack in front.} \\
                                   & \small{The person is raising a backpack with his right hand.} \\
                                   & \small{The person at the front presently has control over the backpack. }\\ \hline
\multirow{2}{*}{\textit{chairwood}} &\small{A person is using the chairwood for sitting.} \\
             (\textit{wooden chair}) & \small{The person is propelling the chairwood on the ground.} \\
                                   & \small{Someone is hoisting a chairwood by his left hand.}\\ \hline
\multirow{3}{*}{\textit{tablesquare}} &\small{A person is lifting the tablesquare, utilizing his left hand.} \\
               (\textit{square table})  & \small{Someone is clutching onto a tablesquare from the front.} \\
                                   & \small{An individual is moving the tablesquare back and forth.}\\ \hline
\multirow{3}{*}{\textit{boxlong}} &\small{A person is gripping the boxlong from the front.} \\
                (\textit{long box})  & \small{A person is raising the boxlong using his left hand.} \\
                                   & \small{Someone hoists the boxlong with his left hand.}\\ \hline
\multirow{3}{*}{\textit{toolbox}} &\small{Someone is grasping the toolbox upfront.} \\
                                  &  \small{The person has a firm hold on the toolbox with his right hand.} \\
                                   & \small{A person is gripping the toolbox with his left hand.}\\  \hline
\multirow{3}{*}{\textit{yogaball}} &\small{A person is shifting a yogaball back and forth on the floor using his hands.} \\
                                   & \small{The person is occupying a yogaball.} \\
                                   & \small{A person is employing an yogaball to engage in an upper body game.}\\ 
\bottomrule
\end{tabular}
}
\end{center}
\end{table}

\myparagraph{Examples of Annotated Textual Descriptions.} 
In \Tref{tab:texts}, we showcase a selection of our annotated textual descriptions for the BEHAVE dataset~\citep{bhatnagar22behave}.

\begin{figure}[t]
    \centering
    \includegraphics[width=1.0\linewidth]{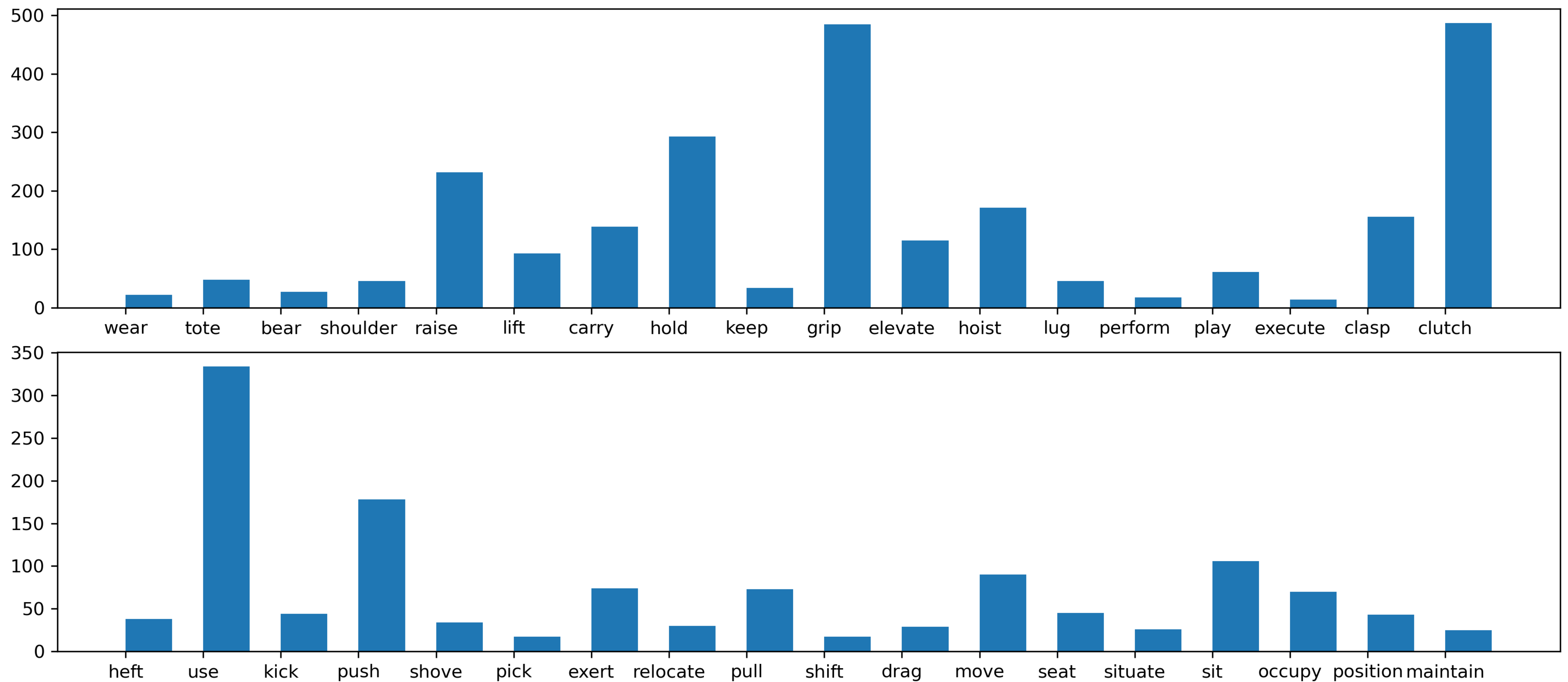}
    \caption{\textbf{Analysis of word frequency} We count the occurrences of each interaction verb from all text descriptions to illustrate their respective frequencies.
    }
    \label{fig:word_freq}
\end{figure}

\myparagraph{Analysis of Annotated Textual Descriptions.} 
\new{All text descriptions encompass 36 distinct interaction verbs associated with 20 different objects. \Fref{fig:word_freq} illustrates the frequency of each verb, indicating their respective occurrences.}

\myparagraph{Affordance Data.} 
Our affordance data includes  8-dimensional human contact labels and object contact points. We employ \textit{chamfer distance} to measure the distance between all human body joints and object surface points. Following a predefined distance threshold $\gamma=0.12$, we identify the 8 contact points on the object surface corresponding to the 8 primary human body joints. Subsequently, we derive the human contact labels by encoding the indexes of contact joints into an 8-dimensional vector represented by binary values. 

\vspace{-0.8em}
\section{Additional Details of OMOMO Dataset}
\label{supp:omomo}
The OMOMO dataset comprises data captured for a total of 15 objects. Adhering to their official split strategy depicted in~\cite{li2023object}(Figure 5), we allocate 10 objects for training and 5 objects for testing. This split allows us to further evaluate the model's generalization ability to new objects. Notably, the OMOMO dataset itself provides text annotation, and we use GPT-3.5 to add subjects to it and embellish it appropriately. For affordance data, we preprocess it the same way we handle BEHAVE.

\section{Common Questions}
\label{supp:questions}
\myparagraph{Why use Skeletal Pose Representation rather than SMPL parameters?}
Most state-of-the-art text-to-motion methods adopt the skeletal pose representation proposed by~\citet{Guo_2022_CVPR}, demonstrating excellent performance and stability. While some works~\citep{azadi2023make} argue that SMPL parameters~\citep{SMPL:2015} contains shape and global information, it does not generate as smooth motions as skeletal-based approaches. Consequently, we adopt the skeletal pose representation and aim to leverage strong pose priors from the pretrained text-to-motion model~\citep{tevet2023human} to ensure the authenticity of generated human motion.

\myparagraph{Can we handle multi-phase interactions between humans and objects?}
Due to the lack of fine-grained textural descriptions in the current 3D HOI dataset, we primarily consider only one interaction phase. However, we have found that an LLM can still reason well for multiple phases given a template such as: \textit{You will be given a sentence that describes an interaction between a person and an object across multiple phases. Your task is to divide the interaction into phases based on the state of the object and determine the state for each phase. If the object is being moved by the person during a phase, output the number 0. If the object remains stationary during a phase, output the number 1.} 

For example, given the text description: \textit{The box is on the ground. A person is picking up the box and holding it forward, then putting the box towards the table. The box is on the table" The result from GPT-3.5-turbo: "Phase 1: The box is on the ground - State: 1 (stationary); Phase 2: The person is picking up the box and holding it forward - State: 0 (moved); Phase 3: The person is putting the box towards the table - State: 0 (moved); Phase 4: The box is on the table - State: 1 (stationary).} We will address the generation of multiple phases of 3D HOI in future work.

\myparagraph{Can we generate hand motion with articulated fingers?}
The BEHAVE and OMOMO datasets do not capture and provide raw hand parameters, despite utilizing SMPLH and SMPLX models to fit human body meshes for rendering. Consequently, in this paper, we focus solely on whole-body human motion, excluding articulated hand and finger movements.

\myparagraph{Why do we use large language models (LLMs) to predict object state based on the input description?}
We aim to leverage LLMs for inferring object states, and our results demonstrate that they perform efficiently and effectively. As shown in the~\Tref{table:LLMs}, we evaluated the performance of object state prediction with GPT-3.5-turbo~\citep{chatgpt} and obtained an average precision of 95.6\% on the validation set, with an average response time of 0.518 seconds. The results suggest that GPT-3.5-turbo is sufficiently accurate without adding significant overhead.  We also evaluated the prediction performance using other LLMs, including Gemini-1.5-Pro-Exp-0801~\citep{reid2024gemini} (99.4\%, 0.259s), Gemma-2-27B~\citep{team2024gemma} (98.6\%, 0.522s), and LLaMA-2-13B~\citep{touvron2023llama} (94.4\%, 0.521s), the latter two being publicly available.

\begin{table}[t]
\centering
\resizebox{0.8\linewidth}{!}{
\begin{tabular}{l|c|c}
    \toprule
    & Acc (\%) $\uparrow$ & Time (s) $\downarrow$\\ 
     \midrule
   GPT-3.5 & 95.6  & 0.518\\
   Gemini-1.5-Pro-Exp-0801 & 99.4 & 0.259\\
    Gemma-2-27B & 98.6 & 0.522\\
    LLaMA-2-13B & 99.4 & 0.259\\
   APDM + MLP  & 79.5 & 2.420 \\
    \bottomrule
\end{tabular}
}
\caption{LLMs' inference accuracy (Acc) and average inference time (Time) on object state prediction. }
\label{table:LLMs}
\end{table}

To further validate the effectiveness of the LLM module, we modified the APDM module by adding an MLP head to predict the object status. The newly added MLP takes in the features consisting of object geometry information and CLIP embeddings. We used an MSE loss. We got average precision 79.5\% and average time 2.42s for this design on the validation set, which is significantly worse than the results of GPT-3.5-turbo (95.6\%, 0.518s), Gemma-2-27b (98.6\%, 0.522s), Gemini-1.5-Pro-Exp-0801 (99.4\%, 0.259s) and LLaMA-2-13B (4.4\%, 0.521s).

In future work, we believe the LLM can play a more important role in 3D HOI, e.g. providing high-level instruction for more complex human-object interactions, and our initial use of the LLM offers insights into its potential applications and how it can be effectively utilized.

\vspace{-0.8em}
\section{Supplementary Video}
\label{supp:video}
Beyond the qualitative results presented in the main paper, our supplementary materials offer comprehensive demos that provide an in-depth visualization of our task, further showcasing the effectiveness of our approach.

In these demonstrations, we highlight the better performance of our method, HOI-Diff, in producing diverse and realistic 3D HOIs while maintaining adherence to physical validity. Notably, the visualizations show that HOI-Diff consistently generates smooth, vivid interactions, accurately capturing human-object contacts.

Additionally, we present the visual ablation results and emphasize the significance and effectiveness of our affordance-guided interaction correction, underscoring its substantial impact on improving the overall performance and quality of the generated 3D HOIs.

\vspace{-0.8em}
\section{Limitations}
\label{supp:limit}
The existing datasets for 3D HOIs are limited in terms of action and motion diversity, posing a challenge for synthesizing long-term interactions in our task. 
Furthermore, the effectiveness of our model's interaction correction component is contingent on the precision of affordance estimation. Despite simplifying this task, achieving accurate affordance estimation remains a significant challenge, impacting the overall performance of our model. A promising direction for future research involves integrating a sophisticated affordance model pre-trained on an extensive 3D object dataset, along with text prompts. Such an advancement could significantly enhance the realism and accuracy of human-object contact in our model, leading to more natural and precise HOIs synthesis.

\vspace{-0.8em}
\section{Social Impacts}
\label{supp:social_impact}
On the positive side, it may offers the research community valuable insights into understanding human behaviors. On the negative side, it remains uncertain whether individuals can be identified solely based on their poses and movements. However, compared to traditional input images of people, this method poses a lower risk of invading personal privacy.
\FloatBarrier

\end{document}